\documentclass[10pt,twocolumn,letterpaper]{article}

\usepackage{wacv}              

\definecolor{wacvblue}{rgb}{0.21,0.49,0.74}
\usepackage[pagebackref,breaklinks,colorlinks,allcolors=wacvblue]{hyperref}

\usepackage{orcidlink}

\usepackage{booktabs}       
\usepackage{amsfonts}       
\usepackage{wrapfig}        
\usepackage{nicefrac}       
\usepackage{microtype}      
\usepackage{xcolor}         
\usepackage{caption}
\usepackage{algpseudocode}
\usepackage{algorithm}
\usepackage{tikz}
\usetikzlibrary{shapes, arrows}
\usepackage{tikz-3dplot}
\usetikzlibrary{positioning, shapes.geometric}
\usepackage{xcolor}
\usepackage{tcolorbox}
\usepackage{lipsum}
\usepackage{balance}
\usepackage{makecell}
\usepackage[]{mdframed}
\usepackage{multirow}
\usepackage{fontawesome}
\usepackage{subcaption}
\usepackage{wrapfig}
\usepackage{graphicx}
\definecolor{pastelgreen}{rgb}{0.75, 0.95, 0.75}
\definecolor{pastelblue}{rgb}{0.75, 0.85, 0.95}
\usepackage[table]{xcolor}
\usepackage{tikz}
\usepackage{nicematrix}
\usepackage{rotating}
\usepackage{yfonts}
\usetikzlibrary{positioning, shapes.geometric}
\usetikzlibrary{arrows.meta}
\usepackage{amsmath, amssymb}
\usepackage{mathrsfs}
\usepackage{mathtools}
\usepackage{ulem}

\definecolor{cvprblue}{rgb}{0.21,0.49,0.74}
\definecolor{codebrown}{rgb}{0.37,0.0,0.0}
\definecolor{codecomment}{rgb}{0.418, 0.424, 0.760}
\definecolor{codegreen}{rgb}{0.024, 0.601, 0.012}
\definecolor{bracketcolor1}{rgb}{0.540, 0.189, 0.511}
\definecolor{bracketcolor2}{rgb}{0.197, 0.730, 0.588}
\newcommand{\formattedparagraph}[1]{\noindent \textbf{#1}}
\newcommand{\nrsfm}{{{NRS\textit{f}M}}}

\newcommand{\revisedone}[1]{\textcolor{black}{#1}}

\def\wacvPaperID{839} 
\def\confName{WACV}
\def\confYear{2027}

\title{No Cones, No Duals: A Fast and Scalable Solver for \nrsfm}

\author{Suryansh Kumar\\
Visual and Spatial AI Lab, VCCM Section  \\ College of PVFA, Department of ECEN, Department of CSCE, \\ Texas A\&M University, College Station, Texas, USA
}

\begin{document}
\maketitle
\begin{abstract}
Non-rigid structure from motion (NRSfM) aims to recover camera motion and time-varying 3D shape from tracked 2D image measurements. Recent high-performing methods show that shape-space factorization remains highly effective for this task. These methods often use costly solvers for two key steps. First, a semidefinite programming (SDP) cone solver to perform Euclidean correction. Next, an alternating direction method of multipliers (ADMM) solver to recover the low-rank shape. These steps often scale poorly to long sequences and dense point tracks. We propose a fast, scalable NRSfM solver that avoids these costly steps. For rotation estimation, we introduce a differentiable Euclidean correction that directly optimizes small factors for each sequence. A row-polar projection then maps each estimated camera block to the closest row-orthonormal matrix. For shape estimation, we introduce an unrolled spectral-proximal solver. It combines reprojection-gradient steps with weighted singular-value shrinkage in the reshuffled shape domain. This solver further uses FISTA-style restart and continuation. The spectral shrinkage operator itself is classical, but we embed it in a fixed differentiable solver, while preserving NRSfM's geometric constraints. The proposed approach is category-agnostic, requires no external training data, and maps naturally to modern accelerator-friendly linear algebra primitives. Across sparse, dense, noisy, missing-data, and long-sequence benchmarks, our solver achieves substantial speedups with competitive reconstruction accuracy, including 7$\times$--1000$\times$ runtime improvements on dense sequences with 28,887 tracked points.
\end{abstract}
    
\section{Introduction}\label{sec:intro}
This paper concerns a matrix-factorization approach to solve Non-Rigid Structure from Motion (\nrsfm). It is a long-standing inverse problem in geometric computer vision \cite{1885-164278}. It seeks to recover camera motion and the 3D shape of a dynamically deforming object from tracked 2D image measurements. \nrsfm ~applications include medical imaging \cite{garg2013dense,sengupta2021colonoscopic, kumar2019jumping}, deformable and soft robotic perception \cite{lamarca2020defslam}, motion capture \cite{akhter2009nonrigid, kumar2016multi, kumar2019superpixel, kumar2017monocular}, and visual effects \cite{parashar20193dvfx, grasshof2022tensor}.

\nrsfm ~factorization remains difficult because many plausible motion-shape decompositions can often explain the same 2D tracks, and both the camera motion and the deforming 3D shape are unknown \cite{kumar2020non, kumar2020dense}. Prior work also studies matrix factorization approaches under weak-perspective \cite{gotardo2011computing, torresani2008nonrigid, kong2016prior} and perspective camera models \cite{lee2013procrustean, dai2013projective}. This paper focuses on the classical orthographic setting with tracked image points. The measurement matrix is modeled as the product of a block-orthographic camera matrix and a shape matrix. Each camera block has orthonormal rows, while the deforming shape is assumed to span a low-rank space.

\begin{figure}[t]
    \centering
    \includegraphics[width=1.0\linewidth]{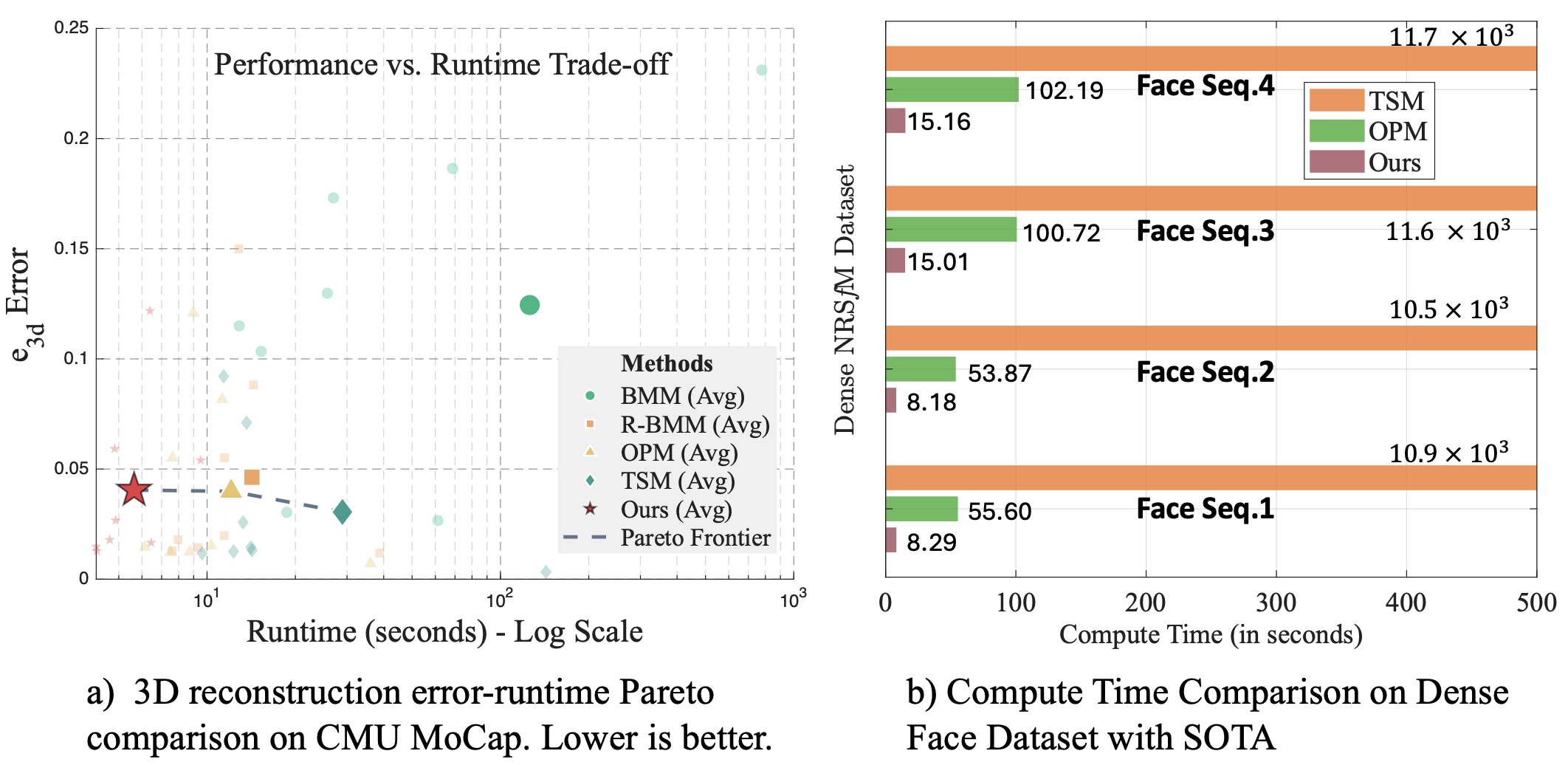}
    \caption{\small Accuracy-runtime and dense-sequence behavior. (a) Accuracy-runtime Pareto comparison on CMU MoCap \cite{akhter2009nonrigid}. Smaller, faded markers show individual sequences, while larger markers show method averages (lower left is better). The proposed solver achieves the fastest runtime while retaining competitive reconstruction accuracy relative to recent shape-space baselines. (b) On dense sequences \cite{garg2013dense} with $28,887$ tracks, the proposed solver achieves large runtime gains over recent shape-space baselines such as TSM \cite{shi2024non} and OPM \cite{kumar2022organic}, while retaining competitive accuracy.}
    \label{fig:intro_figure}
\end{figure}

Factorized \nrsfm ~representations include trajectory-space \cite{akhter2011trajectory}, shape-space \cite{bregler2000recovering}, and shape-trajectory models \cite{kumar2017spatio}. Other approaches use force-based models \cite{agudo2017force}, unions of subspaces \cite{zhu2014complex, kumar2016multi, kumar2017spatio}, unions of piecewise curve subspaces \cite{agudo2021piecewise}, or probabilistic point-trajectory models \cite{agudo2018scalable}. Among these, trajectory-space and shape-space methods are particularly influential and widely used. Akhter et al. \cite{akhter2009nonrigid,akhter2011trajectory} pioneered the \textit{trajectory-space} view. It represents each 3D point's motion with a small set of temporal basis functions. Naturally, the popular choice here is the discrete cosine transform (DCT) basis. This representation is fast to compute. However, a small fixed temporal basis may poorly capture complex or high-frequency deformations \cite{park20153d,zhu2013convolutional}. This can limit reconstruction accuracy \cite{jensen2018benchmark}. A dual view to this problem is shape-space approaches. Bregler et al. \cite{bregler2000recovering} introduced this in early 2000. Dai et al.~\cite{dai2014simple} and Kumar and Van Gool \cite{kumar2022organic} later advanced this approach and showed outstanding results. These methods model shape at each frame and impose a low-rank constraint across time. With careful Euclidean correction, they often achieve higher reconstruction accuracy than trajectory-space methods \cite{jensen2018benchmark,kumar2022organic}. Nevertheless, this accuracy comes at a computational price.

Recent shape-space methods \cite{kumar2022organic,shi2024non} achieve excellent reconstruction results but rely on two costly solver components. First, Euclidean correction estimates or camera rotation initialization. This step often uses semidefinite programming (SDP) or related global optimization over Gram matrices \cite{dai2014simple,kumar2022organic,shi2024non}. Second, a low-rank recovery step estimates shape with the rotations fixed. It typically uses nuclear or weighted nuclear penalties and ADMM-style updates \cite{boyd2011distributed,kumar2022organic,shi2024non}. SDP cone solvers used for rotation estimation are difficult to parallelize efficiently. Meanwhile, ADMM introduces auxiliary variables, dual updates, and large subproblem solves. Moreover, it is sensitive to the method's stopping criteria. These costs become a practical bottleneck as video grow longer or point tracks become denser.

We ask whether a fast and scalable solver can preserve the algebraic structure behind accurate shape-space \nrsfm. Our answer is the \textbf{No Cones, No Duals} solver. No Cones means that rotation recovery avoids SDP cone solvers. It optimizes a small factor parameterization directly rather than solving for a positive semidefinite Gram matrix. No Duals means that shape recovery avoids ADMM dual variables and large shape-subproblem solves. It replaces alternating primal, auxiliary, and dual updates with a fixed unrolled spectral-proximal loop. The solver retains the algebraic and geometric structure of classical \nrsfm. It expresses this structure through differentiable linear algebra operations suited to efficient computation on accelerators.

\formattedparagraph{Rotation Solver.} Classical Euclidean correction solves for Gram matrices satisfying positive semidefiniteness, rank, and orthonormality constraints \cite{dai2014simple}. We instead parameterize each triplet-side Gram matrix to enforce positive semidefiniteness and rank at most 3, while avoiding the explicit SDP cone solver. The factors are optimized using first-order gradient descent with the introduced rotation loss, which measures row-norm equality and row orthogonality after multiplication by the factorized motion matrix. The resulting camera blocks are then mapped to the closest row-orthonormal matrices by a row-polar projection. Thus, the factor-side parameterization, orthonormality loss, and row-polar projection together replace SDP-based Euclidean correction.

\formattedparagraph{Shape Solver.} We optimize the shape objective function via an ADMM-free spectral-proximal solver. We embed the spectral-proximal shrinkage operator in a fixed differentiable \nrsfm ~solver layer that preserves the low-rank shape prior while removing ADMM's auxiliary and dual variables\footnote{We emphasize that weighted singular-value shrinkage and proximal-gradient methods are classical tools and not our claimed novelty.}. Each layer performs a reprojection-gradient step, reshapes the estimate into the standard shape-space matrix, applies weighted singular-value shrinkage, reshapes back, and uses FISTA-style restart and continuation for stable progress. This gives predictable runtime for a fixed number of layers (deterministic), while retaining the option to run more layers when accuracy is preferred over speed.

\begin{table}[t]
\centering
\small
\setlength{\tabcolsep}{3pt}
\renewcommand{\arraystretch}{1.05}
\begin{NiceTabular}{p{0.25\linewidth}| p{0.34\linewidth}|p{0.34\linewidth}}
\toprule
\Block[tikz={top color=orange!25}]{*-1}{}\textbf{Stage} & \Block[tikz={top color=lightgray!25}]{*-1}{}\textbf{Current Solvers} &  \Block[tikz={top color=OliveGreen!25}]{*-1}{} \textbf{Ours} \\
\midrule
Rotation
& SDP/cone solver for Gram recovery
& Factor-side $\mathbf{Y}_k$ + row-polar projection \\
\hline
Shape
& ADMM with auxiliary/dual variables
& Unrolled layer, WSVT shrinkage, restart \\
\hline
Camera
& Orthographic camera constraints
& row-polar camera projection \\
\hline
Computation
& Cone, ADMM subproblem machinery
& Accelerator-friendly solver \\
\bottomrule
\end{NiceTabular}
\caption{No-cone, no-dual positioning relative to recent high-performing shape-space \nrsfm ~factorization pipelines.
}
\label{tab:classical_vs_ours}
\end{table}

Our paper's positioning also differs from supervised deep learning-based \nrsfm ~methods
\cite{kong2019deep,novotny2019c3dpo,park2020procrustean,sidhu2020neural,deng2025deep}. \revisedone{Supervised deep \nrsfm ~methods learn category-level priors from training sequences and are therefore complementary to classical factorization. Since our goal is a category-agnostic per-sequence solver, we focus our comparison on factorization-based \nrsfm ~methods, while reporting additional comparisons to learning-based approaches}. Our solver remains category-agnostic and sequence-specific. It uses differentiable optimization modules, but it does not learn an external object prior. Meanwhile, it is also useful to distinguish our shape solver from dense variational \nrsfm, especially Garg et al. \cite{garg2013dense}, as it was an earlier attempt at an accelerator-friendly approach. Garg et al. \cite{garg2013dense} introduced a dense surface-reconstruction formulation that combines trace-norm regularization with edge-preserving total-variation spatial smoothness on an image-grid surface, and alternates rotation refinement with shape optimization through auxiliary variables, primal-dual TV updates, linear-system solves, and soft-impute-style shrinkage. \revisedone{On the contrary, our method estimates rotations through factor-side Euclidean correction and then applies a fixed no-dual shape solver. The solver, sequential approach pose and shape estimation, and the optimization procedure are different.}

Across sparse and dense benchmarks, our solver gives a better accuracy-runtime tradeoff. Note that it is not designed to dominate every specialized prior in 3D reconstruction error metric, but it consistently improves scalability while remaining competitive in accuracy. In particular, on dense sequences it provides large speedups over recent shape-space baselines while avoiding the SDP and ADMM machinery. Figure \ref{fig:intro_figure} and Table \ref{tab:classical_vs_ours} summarize the proposed solver's performance and positioning, respectively. Overall, this paper makes the following contributions:
\begin{itemize}
    \item We introduce an SDP-free factor-side Euclidean correction for \nrsfm ~rotation estimation. The method optimizes small per-sequence factors and obtains valid orthographic camera blocks via row-polar projection, avoiding explicit cone solvers for Gram recovery.
    
    \item We introduce an ADMM-free unrolled spectral-proximal solver for low-rank shape recovery. Our approach removes ADMM dual variables and large shape-subproblem solves.
    
    \item We put forward a fast and scalable approach, which is category-agnostic, requires no external training data, and maps to accelerator-friendly linear algebra primitives.
    
    \item We evaluate the solver across sparse, dense, missing-data, noisy, and long-sequence benchmarks, showing substantial runtime gains with competitive reconstruction accuracy.
\end{itemize}

\smallskip
\noindent
\revisedone{The central contribution is not a new low-rank prior or a new spectral shrinkage operator or the first scalable or gradient-based \nrsfm~method}, it is a new solver pathway for shape-space \nrsfm ~that replaces the prevailing conic and dual-variable solver machinery.
\section{Preliminaries}\label{sec:preliminaries}

\formattedparagraph{Problem setup.}
We are given a mean-centered measurement matrix
$\mathbf{W}\in\mathbb{R}^{2F\times P}$ containing the tracked 2D coordinates of $P$ points across $F$ frames \cite{bregler2000recovering, tomasi1992shape}. Each frame contributes two rows corresponding to the image $x$--$y$ coordinates of all tracked points. Under the orthographic
camera model, the standard \nrsfm ~factorization writes
$\mathbf{W}\approx \mathbf{R}\mathbf{S}$, where $\mathbf{R}\in\mathbb{R}^{2F\times 3F}$ is block diagonal and
$\mathbf{S}\in\mathbb{R}^{3F\times P}$ stacks the time-varying 3D shapes.
The unknowns are constrained in two different ways.  \textit{(i)} each camera block must have orthonormal rows, i.e., $\mathbf{R}_i\mathbf{R}_i^T=\mathbf{I}_2$, and \textit{(ii)} the deforming 3D shape must admit a low-dimensional representation over time~\cite{akhter2009nonrigid,dai2012simple}. The difficulty of \nrsfm ~comes from satisfying these constraints while explaining the same 2D observations through the bilinear product $\mathbf{R}\mathbf{S}$.

\smallskip
\formattedparagraph{$\bullet$~Rotation Estimation.}
A standard route to camera recovery begins with a $\mathrm{rank}$-$3K$ factorization of the mean-centered measurement matrix,
\[
    \mathbf{W}\approx \mathrm{\hat{\Pi}}\hat{\mathbf{B}},
    ~~\textrm{where} ~\mathrm{\hat{\Pi}}\in\mathbb{R}^{2F\times 3K},
    ~~\hat{\mathbf{B}}\in\mathbb{R}^{3K\times P},
\]
and $K$ is the chosen shape-space rank. This factorization is not unique and for any invertible matrix $\mathrm{\Phi}\in\mathbb{R}^{3K\times 3K}$,
\[\mathbf{W}
    = (\mathrm{\hat{\Pi}}\mathrm{\Phi})
    (\mathrm{\Phi}^{-1}\hat{\mathbf{B}})
    := \mathrm{\Pi}\mathbf{B}.
\]
The role of Euclidean correction is to determine an upgrade
$\mathrm{\Phi}$ such that the upgraded motion matrix contains valid
orthographic camera blocks. Let $\mathrm{\Phi}=[\mathrm{\Phi}_1,\mathrm{\Phi}_2,\ldots,\mathrm{\Phi}_K]$,
where each triplet $\mathrm{\Phi}_k\in\mathbb{R}^{3K\times3}$ corresponds to one shape basis component. Classical shape-space methods introduce the triplet Gram matrix $\mathrm{\Psi}_k:=\mathrm{\Phi}_k\mathrm{\Phi}_k^T.$
By construction, $\mathrm{\Psi}_k\succeq 0$ and $\mathrm{rank}(\mathrm{\Psi}_k)\leq 3$, with rank three desired in the
nondegenerate case. Let $\hat{\pi}_{2i-1}$ and $\hat{\pi}_{2i}$ denote the
$(2i-1)$-th and $2i$-th rows of $\mathrm{\hat{\Pi}}$, respectively. If the
upgraded triplet agrees with a scaled camera block, then
\begin{equation}
    \hat{\pi}_{2i-1:2i}\mathrm{\Phi}_k = c_{ik}\mathbf{R}_i,
    \label{eq:rotation_gram_constraint}
\end{equation}
for some scalar coefficient $c_{ik}$. Since the two rows of $\mathbf{R}_i$
are orthonormal, the two rows of $\hat{\pi}_{2i-1:2i}\mathrm{\Phi}_k$ must have equal squared norm and zero inner product. Written in terms of $\mathrm{\Psi}_k$, this gives the standard Euclidean-correction constraints
\begin{equation}
\begin{aligned}
    \hat{\pi}_{2i-1}\mathrm{\Psi}_{k}\hat{\pi}_{2i-1}^{T}
    &=
    \hat{\pi}_{2i}\mathrm{\Psi}_{k}\hat{\pi}_{2i}^{T},  
    ~~\hat{\pi}_{2i-1}\mathrm{\Psi}_{k}\hat{\pi}_{2i}^{T}
    &=0 .
\end{aligned}
\label{eq:orthonormality_contraint}
\end{equation}
Solving Eq.~\eqref{eq:orthonormality_contraint} together with
$\mathrm{\Psi}_{k}\succeq 0$ and a low-rank recovery condition gives a
principled metric upgrade~\cite{dai2012simple,dai2014simple}. However, this
formulation operates over positive semidefinite matrix variables and typically requires an SDP or related cone-solver machinery \cite{dai2014simple, shi2024non}. This is the rotation-side bottleneck that our approach addresses.

\smallskip
\formattedparagraph{$\bullet$~Shape Estimation.}
Once the camera blocks $\{\mathbf{R}_i\}_{i=1}^{F}$ are known, the remaining
problem is to estimate the time-varying 3D shape. Let
$\mathbf{w}_{ij}\in\mathbb{R}^{2}$ be the observed 2D point $j$ in frame $i$, and let $\mathbf{s}_{ij}\in\mathbb{R}^{3}$ be the corresponding 3D point in $\mathbf{S}_i$. The reprojection term alone is insufficient, so shape-space \nrsfm ~imposes a low-rank prior on the sequence of shapes. It is common to use the reshuffled shape matrix $
\mathbf{S}^{\sharp}=g(\mathbf{S})\in\mathbb{R}^{F\times 3P},$ whose $i$-th row contains the flattened $3\times P$ shape of the $i$-th frame \cite{dai2012simple,iglesias2020accurate,kumar2022organic,shi2024non}.
When the deforming object is described by $K$ basis shapes, $\mathrm{rank}(\mathbf{S}^{\sharp})\leq K$ provides a direct shape-space low-rank constraint. A schematic rank-regularized formulation is
\begin{equation}
\begin{aligned}
    & \displaystyle \min_{\mathbf{S}}
    \frac{1}{2}\sum_{i=1}^{F}\sum_{j=1}^{P}
    \left\|
        \mathbf{w}_{ij}-\mathbf{R}_i\mathbf{s}_{ij}
    \right\|_2^2
    +
    \lambda~\mathrm{rank}(\mathbf{S}^{\sharp}), \\
    & \displaystyle \mathrm{subject ~to:}
    ~\mathbf{S}^{\sharp}=g(\mathbf{S})
\end{aligned}
\label{eq:rank_minimization}
\end{equation}
The exact rank penalty is nonconvex. Practical methods therefore replace it by nuclear, weighted nuclear, or related spectral penalties and solve a split optimization problem, often using ADMM-style auxiliary variables and dual updates \cite{kumar2022organic,shi2024non}.
This ADMM machinery is accurate but introduces additional memory, large shape-subproblems, and stopping-criterion sensitivity, which our approach addresses. 

\section{Proposed Approach}\label{sec:approach}
The proposed solver has two stages, mirroring the two bottlenecks discussed in Sec.~\S \ref{sec:preliminaries}. First, we recover camera rotations without solving an SDP cone program. Second, given the rotations, we recover shape without ADMM dual variables or large shape-subproblem solves. 

\subsection{Rotation Estimation}\label{ssec:rotation-estimation}
The classical Euclidean-correction step estimates Gram matrices
$\mathrm{\Psi}_k$ satisfying positive semidefiniteness, low-rank structure, and the frame-wise orthonormality constraints in Eq.~\eqref{eq:orthonormality_contraint}. This is principled, but it leads to a cone-constrained optimization problem over $\mathrm{\Psi}_k$. Our rotation estimation takes a different approach. We change the optimization variable. Instead of optimizing the Gram matrix directly, we optimize its factor. Let
$ \Theta = \{\mathbf{Y}_k\}_{k=1}^{K}, \mathbf{Y}_k\in\mathbb{R}^{3K\times3},$ 
be the set of learnable per-sequence factor variables. We parameterize
\begin{equation}
    \mathrm{\Psi}_k(\Theta) = \mathbf{Y}_k\mathbf{Y}_k^{T}.
    \label{eq:cone_neuralized}
\end{equation}
This guarantees $\mathrm{\Psi}_k\succeq 0$ and
$\mathrm{rank}(\mathrm{\Psi}_k)\leq 3$ by construction. Full non-degeneracy is encouraged by the scale regularizer below and by the final row-polar projection. Most importantly, Eq.\eqref{eq:cone_neuralized} avoids the need to form and solve the global cone system $\mathbf{A}\mathrm{vec}(\mathrm{\Psi}_k)=\mathbf{0}$ associated with Eq.\eqref{eq:orthonormality_contraint}. The learnable variables are the
small factor tensors $\mathbf{Y}_k$ themselves and no MLP or CNN architecture is introduced.

For each frame $i$ and basis index $k$, we synthesize the two rows
$\mathbf{g}_{1,ik} = \hat{\pi}_{2i-1}\mathbf{Y}_k
    \in\mathbb{R}^{1\times3}, ~~\mathbf{g}_{2,ik} =
    \hat{\pi}_{2i}\mathbf{Y}_k \in\mathbb{R}^{1\times3}.$
Stacking them gives
$ \mathbf{\Omega}_{ik} \triangleq [\mathbf{g}_{1,ik};~\mathbf{g}_{2,ik}] \in\mathbb{R}^{2\times3}.$ If $\mathbf{Y}_k$ matches the true corrective triplet $\mathrm{\Phi}_k$ up to the \nrsfm ~scale coefficient, then
$\mathbf{\Omega}_{ik}=c_{ik}\mathbf{R}_i$ due to Eq.\eqref{eq:rotation_gram_constraint}. Hence the two rows of
$\mathbf{\Omega}_{ik}$ should have equal squared norm and zero inner product. We encode these two Euclidean-correction conditions with the nonnegative orthonormality residual
\begin{equation}
\begin{aligned}
\mathcal{L}_{\mathrm{ortho}}(\Theta) = \frac{1}{FK} \sum_{i=1}^{F} \sum_{k=1}^{K} \Big[&\big(\|\mathbf{g}_{1,ik}\|_2^2-\|\mathbf{g}_{2,ik}\|_2^2\big)^2
\\
& + \big(\mathbf{g}_{1,ik}\cdot\mathbf{g}_{2,ik} \big)^2\Big].
\end{aligned}\label{eq:loss_ortho}
\end{equation}
This is the factor-side counterpart of the two linear constraints in
Eq.~\eqref{eq:orthonormality_contraint}. As a result, instead of enforcing the constraint on $\mathrm{\Psi}_k$ via a cone solver, we evaluate their residuals after multiplication by $\mathbf{Y}_k$. Still, the loss in Eq.\eqref{eq:loss_ortho} can admit a trivial solution $\mathbf{Y}_k=\mathbf{0}$. We therefore add a scale regularizer 
$s_{ik} = 0.5 \left(\|\mathbf{g}_{1,ik}\|_2^2 + \|\mathbf{g}_{2,ik}\|_2^2\right),$ as \textbf{scale loss}: 
\begin{equation}
\mathcal{L}_{\mathrm{scale}}(\Theta) = \frac{1}{K} \sum_{k=1}^{K}\left(
\frac{1}{F}\sum_{i=1}^{F}s_{ik}-\tau_s\right)^2, ~\tau_s\approx 1.
\label{eq:loss_scale}
\end{equation}
This softly fixes the average row scale while preventing the degenerate condition. To control the scale of the factor variables and optimization stability, we also include a mild trace or Frobenius  regularizer as \textbf{trace loss}:
\begin{equation}
    \begin{aligned}
        \displaystyle \mathcal{L}_\text{trace}(\Theta) = \frac{1}{K} \sum_{k=1}^{K} \textrm{trace}(\mathbf{Y}_{k}\mathbf{Y}_{k}^{T}) = \frac{1}{K} \sum_{k=1}^{K}\|\mathbf{Y}_{k}\|_{F}^2
    \end{aligned}\label{eq:loss_trace_rotation}
\end{equation}
Together, the scale and trace regularizers (Eq.\eqref{eq:loss_scale} and Eq.\eqref{eq:loss_trace_rotation}) discourage degenerate factors and stabilize optimization. The key observation is that evaluating residuals in the measurement space via $\textbf{g}_{1, ik} , ~\textbf{g}_{2, ik}$ is cheaper and numerically more stable. Consequently, the global null space system $A \text{vec}(\Psi_{k}) = \mathbf{0}$ is unnecessary once we parameterize via $\mathbf{Y}_{k}$. Combining all the above loss functions, gives our overall loss function as
\begin{equation}
    \begin{aligned}
    & \mathcal{L}_{\mathrm{rot}}(\Theta) = \mathcal{L}_{\mathrm{ortho}}(\Theta)
    + \textcolor{black}{\lambda_s}\mathcal{L}_{\mathrm{scale}}(\Theta) + \textcolor{black}{\lambda_t}\mathcal{L}_{\mathrm{trace}}(\Theta), \\
    & \Theta^\star=\arg\min_{\Theta}\mathcal{L}_{\mathrm{rot}}(\Theta).
    \end{aligned}\label{eq:overall_rotation_loss}
\end{equation}
All terms in Eq.~\eqref{eq:overall_rotation_loss} are evaluated through
matrix multiplications involving $\hat{\Pi}$ and $\mathbf{Y}_k$, and are therefore simple to optimize with first-order gradient methods. After optimizing $\Theta$, we select the basis triplet with the smallest average orthonormality residual $k^\star = $ 
\begin{equation}
\begin{aligned} \arg\min_{k}\frac{1}{F}
\sum_{i=1}^{F}\Big[&\big(\|\mathbf{g}_{1,ik}\|_2^2-\|\mathbf{g}_{2,ik}\|_2^2 \big)^2 + 
\big(\mathbf{g}_{1,ik}\cdot\mathbf{g}_{2,ik} \big)^2 \Big]
\end{aligned}
\label{eq:rotation_basis}
\end{equation}
The selection of $k^\star$ is a final discrete model-selection step after optimizing the factor variables. For each frame, the selected block
$\mathbf{\Omega}_{ik^\star} =
[\mathbf{g}_{1,ik^\star};~\mathbf{g}_{2,ik^\star}]
\in\mathbb{R}^{2\times3}$
is then mapped to the closest row-orthonormal camera block using
\begin{equation}
    \mathbf{R}_i = (\mathbf{\Omega}_{ik^\star}
    \mathbf{\Omega}_{ik^\star}^{T})^{-1/2}
    \mathbf{\Omega}_{ik^\star}.
    \label{eq:per_frame_rot}
\end{equation}
Note that the unknown scale $c_{ik^\star}$ cancels in this row-polar projection, and the result satisfies $\mathbf{R}_i\mathbf{R}_i^T=\mathbf{I}_2$ whenever $\mathbf{\Omega}_{ik^\star}$ has full row rank. We compute the
$2\times2$ inverse square root using Newton-Schulz iterations
\cite{kenney1992scaling,rubensson2021localized}, which are lightweight and accelerator-friendly. The full motion matrix is finally assembled as
$\mathbf{R}= \mathrm{blkdiag}(\mathbf{R}_1,\ldots,\mathbf{R}_F).$
A derivation showing Eq.\eqref{eq:per_frame_rot} as the closest
row-orthonormal projection is provided in the supplementary Sec.\S \ref{sec:rowpolar}.

\subsection{Shape Estimation}\label{ssec:shape-estimation}
Given $\mathbf{W}$ and recovered
$\mathbf{R}$, we estimate $\mathbf{S}$. We start from the standard re-projection term with low-rank shape objective
\begin{equation}
\begin{aligned}
\min_{\mathbf{S}} \frac{1}{2} \|\mathbf{M}\odot(\mathbf{W}-\mathbf{R}\mathbf{S})\|_F^2 +
\|\mathbf{S}^{\sharp}\|_{*,w}; ~\mathrm{s.t.} ~\mathbf{S}^{\sharp}=g(\mathbf{S}).
\end{aligned}
\label{eq:shape_optimization}
\end{equation}
Here, $\mathbf{M}\in\{0,1\}^{2F \times P}$ is the visibility mask, where
$\mathbf{M}_{ij}=0$ indicates a missing observation, and
$g(\cdot)$ reshapes $\mathbf{S}\in\mathbb{R}^{3F\times P}$ into
$\mathbf{S}^{\sharp}\in\mathbb{R}^{F\times3P}$, whose $i$-th row is the
flattened $3\times P$ shape of frame $i$. The weighted spectral penalty is
$\|\mathbf{S}^{\sharp}\|_{*,w} = \sum_i w_i\sigma_i(\mathbf{S}^{\sharp}),$
where the singular values are sorted in nonincreasing order. Following prior
weighted-nuclear formulations in \nrsfm ~\cite{kumar2020non}, the weights are computed from the singular values of an initial shape $\mathbf{S}^0=\mathrm{pinv}(\mathbf{R})\mathbf{W}$. The objective in Eq.~\eqref{eq:shape_optimization} has a smooth re-projection term and a non-smooth spectral regularizer. More abstractly, we write
\begin{equation}
    J(\mathbf{S}) = f(\mathbf{S})+\mathcal{G}(\mathbf{S}).
    \label{eq:shape_optimization_abstraction}
\end{equation}
where, $f(\mathbf{S}) = 0.5 \|\mathbf{M}\odot(\mathbf{W}-\mathbf{R}\mathbf{S}) \|_F^2 $ is a smooth function while $\mathcal{G}(\mathbf{S}) = \|\mathbf{S}^{\sharp}\|_{*,w}$ is non-smooth function. The gradient of the smooth term is
\begin{equation}
\nabla f(\mathbf{S}) =
\mathbf{R}^{T}\big( \mathbf{M}\odot(\mathbf{R}\mathbf{S}-\mathbf{W})
\big) \in\mathbb{R}^{3F\times P}.
\label{eq:gradient_data_term}
\end{equation}
Now, for $f(\mathbf{S})$ gradient step to be valid and stable, $\nabla f(\mathbf{S})$ must be \textit{Lipschitz continuous}. Because the mask is nonexpansive and $\mathbf{R}$ is block-orthographic, Lipschitz condition will tell us that $ \nabla f$ does not change too abruptly, and geometrically the curvature of $f$ is bounded.  Under orthographic projection, the gradient is {Lipschitz} with constant $L  \le \|\mathbf{R}\|_2^2 \le 1$ ~\cite{lipschitz1877lehrbuch, nesterov2013introductory,beck2017first}, so $L\le 1$. A fixed step size $\alpha \in (0,1/L] ~\textrm{or} ~\alpha \in (0, 1] $ is valid. This provides a valid fixed step size for the smooth reprojection step. Since the weighted spectral penalty is generally nonconvex, we use monotone restart and continuation for stable practical descent. Refer to the supplementary material Sec.\S \ref{sec:lipschitz} for detailed proof and explanations on Lipschitz condition.

\smallskip
\formattedparagraph{\textit{(i)} Unrolled Spectral-Proximal Solver.} 
$\mathcal{G}(\mathbf{S})$ in Eq.\eqref{eq:shape_optimization_abstraction} is a weighted nuclear norm (WNN) term and is non-smooth. Its derivative is undefined at zero singular values. To handle such a non-smooth regularizer, we used proximal operators, which are a standard tool in optimization \cite{parikh2014proximal}. It provides a closed-form spectral shrinkage step that can be inserted directly after the reprojection-gradient update. Thus, a proximal step is the building block that replaces the dual update in ADMM, but in a more direct and accelerator-friendly way.

Given $\mathbf{S}^\sharp \in \mathbb{R}^{F \times 3P}$, with SVD, $ \mathbf{S}^\sharp = \mathbf{U}\,\mathrm{diag}(\{\sigma_i\})\mathbf{V}^T, $ where $\sigma_i$ are the singular values sorted in nonincreasing order. The weighted singular value thresholding (WSVT) operator acts by shrinking each singular value by an amount proportional to its weight $w_i$:
\begin{equation}
\label{eq:wsvt}
\mathcal{S}_{\tau,w}(\mathbf{S}^\sharp) = \mathbf{U}\,\mathrm{diag}\left(\{\max(\sigma_i - \tau w_i, 0)\}_i\right) \mathbf{V}^T.
\end{equation}
That is, large singular values (important deformation modes) survive the shrinkage, while small ones, i.e., spurious deformations or noise, are driven to zero. Following \cite{kumar2022organic}, we use Eq.\eqref{eq:wsvt} ordered weighted shrinkage update. Because the weighted spectral penalty is generally nonconvex for the weights used here, we do not claim global convex optimality of the full objective. We use this operation as a closed-form spectral shrinkage update inside a monotone-restarted fixed-depth unrolled feed-forward solver \cite{monga2021algorithm, xiang2021fista}. The regularizer remains non-smooth; yet, the shrinkage map is differentiable almost everywhere and can be embedded in an autograd-compatible computation graph.

We combine the reprojection-gradient step and the spectral shrinkage update as follows:
(i) a gradient step on $f(\mathbf{S})$, and (ii) a weighted spectral-shrinkage step on $\mathcal{G}(\mathbf{S})$.
Let $\mathbf{S}^{(k)}$ be the current shape iterate, and $\mathbf{T}^{(k)}$ the FISTA auxiliary variable \cite{beck2009fast, xiang2021fista}.  Each iteration performs: \textit{1. Gradient Step}: $~\mathbf{Z}^{(k)\sharp} = g(\mathbf{T}^{(k)}) - \alpha\, g(\nabla f(\mathbf{T}^{(k)}))$, \textit{2. Proximal shrinkage and reshape}: $\mathbf{X}^{(k)\sharp} = \mathcal{S}_{\tau_k,w}(\mathbf{Z}^{(k)\sharp}), ~\mathbf{S}^{(k+1)} = g^{-1}(\mathbf{X}^{(k)\sharp})$, and \textit{3. FISTA momentum \cite{beck2009fast, xiang2021fista}:} $t_{k+1} = \tfrac12\big(1+\sqrt{1+4t_k^2}\big); \beta_k  = \tfrac{t_k-1}{t_{k+1}}, \mathbf{T}^{(k+1)} = \mathbf{S}^{(k+1)} + \beta_k(\mathbf{S}^{(k+1)}-\mathbf{S}^{(k)})$.  Here, $\alpha$ is a step size and $g(\nabla f(\cdot))$ reshapes the gradient into the reshuffled shape $\mathbf{S}^{\sharp}$ dimension. We use a monotone restart if the cost of $J(\mathbf{S})$ increases, i.e., if 
$J(\mathbf{S}^{(k+1)})>J(\mathbf{S}^{(k)})$, we set
$t_{k+1}=1$ and $\mathbf{T}^{(k+1)}=\mathbf{S}^{(k+1)}$.
Stacking a fixed number $N_k$ of these iterations gives a deterministic feed-forward unrolled shape solver with predictable runtime. A detailed computational pipeline is provided in the supplementary material Sec. \S \ref{sec:supp_pipeline}.

\smallskip
\formattedparagraph{\textit{(ii)} Continuation schedule and practical safeguards.} We set the shrinkage thresholds as $\tau_k=\mu/\rho_k$ (where $\mu \approx 1 $) with $\rho_{k+1}=\min(\rho_{\max},\eta\rho_k)$.
Early iterations (small $\rho$) apply strong shrinkage that rapidly enforces low rank, while later iterations (larger $\rho$) relax the shrinkage to fine-tune re-projection accuracy. This plays a role analogous to ADMM penalty continuation, but without dual variables. We preserve the first $N_{\mathrm{pres}}$ singular values by setting the shrinkage amount $\Delta_i=0$ for $i\leq N_{\mathrm{pres}}$, ensuring that the dominant deformation mode is not shrunk. We also cap shrinkage by replacing $\tau_k w_i$ with $\min(\tau_k w_i, \mathrm{cap} \cdot \sigma_i(\mathbf{Z}^{(k)\sharp}))$ to prevent premature collapse of small but meaningful deformation modes. All continuation and safeguard parameters are fixed across experiments unless otherwise stated.

\smallskip
\formattedparagraph{\textit{(iii)} No-dual solver interpretation.}
The shape stage replaces the ADMM split variables and dual updates with a direct feed-forward sequence layer of gradient and spectral-shrinkage operations. Each layer uses only matrix multiplication, masking, reshaping, SVD, scalar shrinkage, and a FISTA-style skip/restart \cite{beck2009fast, xiang2021fista}. Thus, the method preserves the shape's low-rank without ADMM dual variables, large shape-subproblem linear solves, and dependence on the stopping criterion. A complexity comparison with ADMM-based solvers is provided in the supplementary Sec.~\S \ref{sec:complexity}.

\begin{algorithm}[t]
\footnotesize
\caption{Shape Solver}
\begin{algorithmic}[1]
\Require $\mathbf{R}, \mathbf{W}, \mathbf{M}$; step size $\alpha \in (0, 1/L]$; shrinkage scale $\mu$; weights ${w}~\textrm{(non-decreasing)}$; schedule parameters $\rho_0 >0, \eta >1,\rho_{\max}$; unrolling depth $N_{k}$; options: $N_\textrm{pres} = 1$, cap
\Ensure $ \mathbf{S} \in\mathbb{R}^{3F\times P}$
\State Initialize $\mathbf{S}^{(0)}=\mathbf{R}^\dagger \mathbf{W}$, $\mathbf{T}^{(0)}=\mathbf{S}^{(0)}$, $t_0=1$, $\rho=\rho_0$, $\mu = 0.99$, $\textrm{cap} = 0.99$, $\eta = 1.05$, $N_k = 160$
\For{$k=0$ to $N_k-1$}
  \State $\mathbf{G} \gets \mathbf{R}^T(\mathbf{M} \odot(\mathbf{R} \mathbf{T}^{(k)}- \mathbf{W}))$ ~{\color{codegreen} \# gradient in original domain}
  \State $\mathbf{Z}^\sharp \gets g(\mathbf{T}^{(k)}) - \alpha g(\mathbf{G})$ ~~~~~~~~~~~~~~{\color{codegreen} \# S-sharp gradient step}
  \State $[\mathbf{U},\mathbf{\Sigma},\mathbf{V}]\gets \mathrm{svd} (\mathbf{Z}^\sharp)$; $\tau_{k} \leftarrow \mu/\rho_{k}$
    \For{$i=1$ to $\min(F,3P)$}  ~~~~~~~~~~{\color{codegreen} \# weighted shrinkage}
        \State $w_i \leftarrow \mathbf{w}(i)$; $\Delta_i \leftarrow \tau_{k} w_i$;
        \If {$i \le N_{\mathrm{pres}}$} $\Delta_i\leftarrow 0$
        \EndIf
        \If {cap $ > $ 0} $\Delta_i \leftarrow \min\{\Delta_i,\ \mathrm{cap}\cdot \mathbf{\Sigma}_{ii}\}$ 
        \EndIf
        \State $\mathbf{\Sigma}'_{ii}\leftarrow \max\{\mathbf{\Sigma}_{ii}-\Delta_i,\ 0\}$
    \EndFor
  \State $\mathbf{X}^\sharp \gets \mathbf{U} \mathbf{\Sigma}'\mathbf{V}^T$; ~$\mathbf{S}^{(k+1)}\gets g^{-1}(\mathbf{X}^\sharp)$
  \State {\color{codegreen} \# FISTA with monotone restart}
  \State $t_{k+1}=\tfrac12(1+\sqrt{1+4t_k^2})$; $\beta_k=\tfrac{t_k-1}{t_{k+1}}$
  \State $\mathbf{T}^{(k+1)} = \mathbf{S}^{(k+1)}+\beta_k(\mathbf{S}^{(k+1)}-\mathbf{S}^{(k)})$
  \State {\color{codegreen} \# restart momentum if objective increases}
  \If{$J(\mathbf{S}^{(k+1)})>J(\mathbf{S}^{(k)})$} \\
   ~~~~~~~~~~Restart: $t_{k+1}=1$, $\mathbf{T}^{(k+1)} = \mathbf{S}^{(k+1)}$ 
  \EndIf
  \State $\rho_{k+1} \gets \min(\rho_{\max}, \eta \rho_{k})$ ~~~~~~~~{\color{codegreen} \# continuation}
\EndFor
\State \Return $\mathbf{S}^{(N_{k})}$
\end{algorithmic}\label{alg:shapeestimation}
\end{algorithm}
\section{Experiment and Results}\label{sec:experiments}

\formattedparagraph{\textit{(i)} Implementation Details.} We implemented our approach using Python 3.12.11 and PyTorch 2.5.1. We report wall-clock runtimes on an Apple M3 Max unless otherwise noted. Dense sequence experiments requiring CUDA SVD were run on an NVIDIA RTX 6000 Ada GPU. Runtimes are indicative and emphasize scaling trends. When exact same-hardware reruns were not possible due to released-code or backend limitations, we report the available baseline runtimes and interpret them primarily as scaling indicators. \revisedone{Unless otherwise stated, Table~\ref{tab:rotation_error} reports rotation-stage runtime, Table~\ref{tab:shape_error_1} reports shape-stage runtime after rotations are provided, and Table~\ref{tab:dense_nrsfm_result} reports dense shape-space solver runtime. A stage-wise runtime breakdown is provided in the supplement.}

We ran 3,500 steps to minimize the loss function proposed in Eq.\eqref{eq:overall_rotation_loss} with learning rate $10^{-3}$ and $\lambda_s = 1e^{-4}, \lambda_t = 1e^{-6}$ to recover the rotation $\mathbf{R} \in \mathbb{R}^{2F \times 3F}$. We ran $N_k = 160$ fixed unrolling steps of proximal gradient descent (PGD) to recover shape $\mathbf{S}^{3F \times P} $ by optimizing Eq.\eqref{eq:shape_optimization} using the pseudo code outlined in \textbf{Algorithm}~\ref{alg:shapeestimation}. Note $N_k = 160$ is a conservative fixed budget; in most sequences the objective stabilizes earlier. We initialized $\eta = 1.05$, $\rho_0 = 2e^{-3}$, $\rho_{\max} = 1e^{3}$, $N_\textrm{pres} = 1$, and the weight ${w}$ following \cite{kumar2022organic} work. Refer to \textbf{Algorithm}~\ref{alg:shapeestimation} for step-by-step implementation details. Note that the variable $\textrm{cap} = 0.99$ 
prevents premature spectral collapse over iterative steps.

\smallskip
\formattedparagraph{\textit{(ii)} Evaluation Metric.} Evaluations are done on popular motion capture (MoCap) benchmarks \cite{akhter2009nonrigid, torresani2008nonrigid}, \nrsfm ~challenge \cite{jensen2018benchmark}, dense \nrsfm ~dataset from Garg et al. \cite{garg2013dense}, and  recent H3WB \cite{zhu2023h3wb} dataset. We used mean normalized 3D reconstruction error metric $e_{3d} = \frac{1}{F} \sum_{i=1}^{F} \|\mathbf{S}_{i}^\textrm{est} - \mathbf{S}_{i}^\textrm{gt} \|_{F}/{\|\mathbf{S}_{i}^\textrm{gt}\|_{F}}$ to report 3D reconstruction accuracy. Here, $\mathbf{S}_{i}^\textrm{est}$ and $\mathbf{S}_{i}^\textrm{gt}$ denote the estimated and ground-truth 3D shape for frame $i$, respectively. For experiments on \nrsfm ~challenge benchmark \cite{jensen2018benchmark}, we adopt their official evaluation code. The $e_{3d}$ is reported after registering the estimated shape to the ground truth. To assess pose recovery, we report the mean rotation error, defined as $e_{R} = \frac{1}{F} \sum_{i=1}^{F} \| \mathbf{R}_{i}^\textrm{gt} - \mathbf{R}_{i}^\textrm{est} \|_{F}$, where, $\mathbf{R}_{i}^\textrm{est}$ and  $\mathbf{R}_{i}^\textrm{gt}$ represent the  estimated and ground-truth rotation matrices for frame $i$, respectively.

\subsection{Dataset and Results}
\formattedparagraph{\textit{(i)} MoCap Benchmark.} This benchmark introduced in \cite{akhter2009nonrigid, torresani2008nonrigid} is now a default evaluation suite for \nrsfm ~algorithms. It comprises eight \revisedone{standard} sequences: Drink (1102, 41), Pickup (357, 41), Yoga (307, 41), Stretch (370, 41), Dance (264, 75), Walking (260, 55), Face (316, 40), and Shark (240, 91). The first four sequences \revisedone{have} both ground-truth (GT) rotation and 3D shape, whereas the \revisedone{remaining sequence have} only GT 3D shape for evaluation. The last three sequences were introduced in \cite{torresani2008nonrigid}. The tuples augmented with each dataset represent the number of frames and points, i.e., $(F, P)$. Shark is synthetic, while the remaining follow the MoCap protocol used in prior \nrsfm ~evaluations.

\begin{table}[t]
\centering
\resizebox{1.0\linewidth}{!}{
\begin{tabular}{c|c|c|c|c|c|c}
\hline
Data & $(F, P, K)$ & BMM \cite{dai2014simple}   & R-BMM \cite{kumar2020non} & OPM \cite{kumar2022organic} & TSM \cite{shi2024non} \((\mathbf{R}_{\mathrm{init}} /   ~\mathbf{R}_{\mathrm{ref}})\) & Ours \\ \hline
Drink     & \makecell{(1102, 41) \\ $K = 11$}   & \makecell{0.0086 \\ (10.01)} & \makecell{0.0240 \\ (108.91)}   & \makecell{0.0096 \\ (143.54)} & \makecell{0.0086/ 1.3950 \\ (88.94)} & \makecell{\textbf{0.0060} \\ \textcolor{codebrown}{\textbf{(8.51)}}}  \\ \hline
Pickup    & \makecell{(357, 41) \\ $K = 12$}   & \makecell{0.1210 \\ (09.68)}    &  \makecell{0.1271 \\ (96.24)}   & \makecell{0.1144 \\ (98.87)} & \makecell{0.1210/ 1.3197 \\ (14.79)}& \makecell{\textbf{0.1082} \\ \textcolor{codebrown}{\textbf{(7.79)}}}  \\ \hline
Yoga      & \makecell{(307, 41) \\ $K = 11$}   & \makecell{0.0945 \\ (08.08)}   & \makecell{0.0944 \\ (70.08)}  & \makecell{0.0972  \\ (76.88)} & \makecell{0.0945/ 0.1114 \\ (10.48)} & \makecell{0.0996 \\ \textcolor{codebrown}{\textbf{(7.73)}}} \\ \hline
Stretch   & \makecell{(370, 41) \\ $K = 13$}  & \makecell{0.0682 \\ (08.45)}   &  \makecell{0.0682 \\ (73.32)}   & \makecell{0.0688 \\ (115.06)} & \makecell{0.0682/ 1.4116 \\ (19.40)} & \makecell{\textbf{0.0647} \\ \textcolor{codebrown}{\textbf{(7.83)}}} \\ \hline
\end{tabular}
}
\caption{\scriptsize $e_{R}$ and \revisedone{rotation-stage runtime} comparison with baselines. All methods are re-simulated for given $K$. \revisedone{TSM uses two rotation estimates, i.e., BMM initialization \(\mathbf{R}_{\mathrm{init}}\) and the refined rotation \(\mathbf{R}_{\mathrm{ref}}\) after its refinement module}. Our approach is considerably faster while achieving competitive or better rotation accuracy. The number in parentheses shows compute time in seconds.}\label{tab:rotation_error}
\end{table}

\begin{figure}[t]
    \centering
    \includegraphics[width=0.86\linewidth]{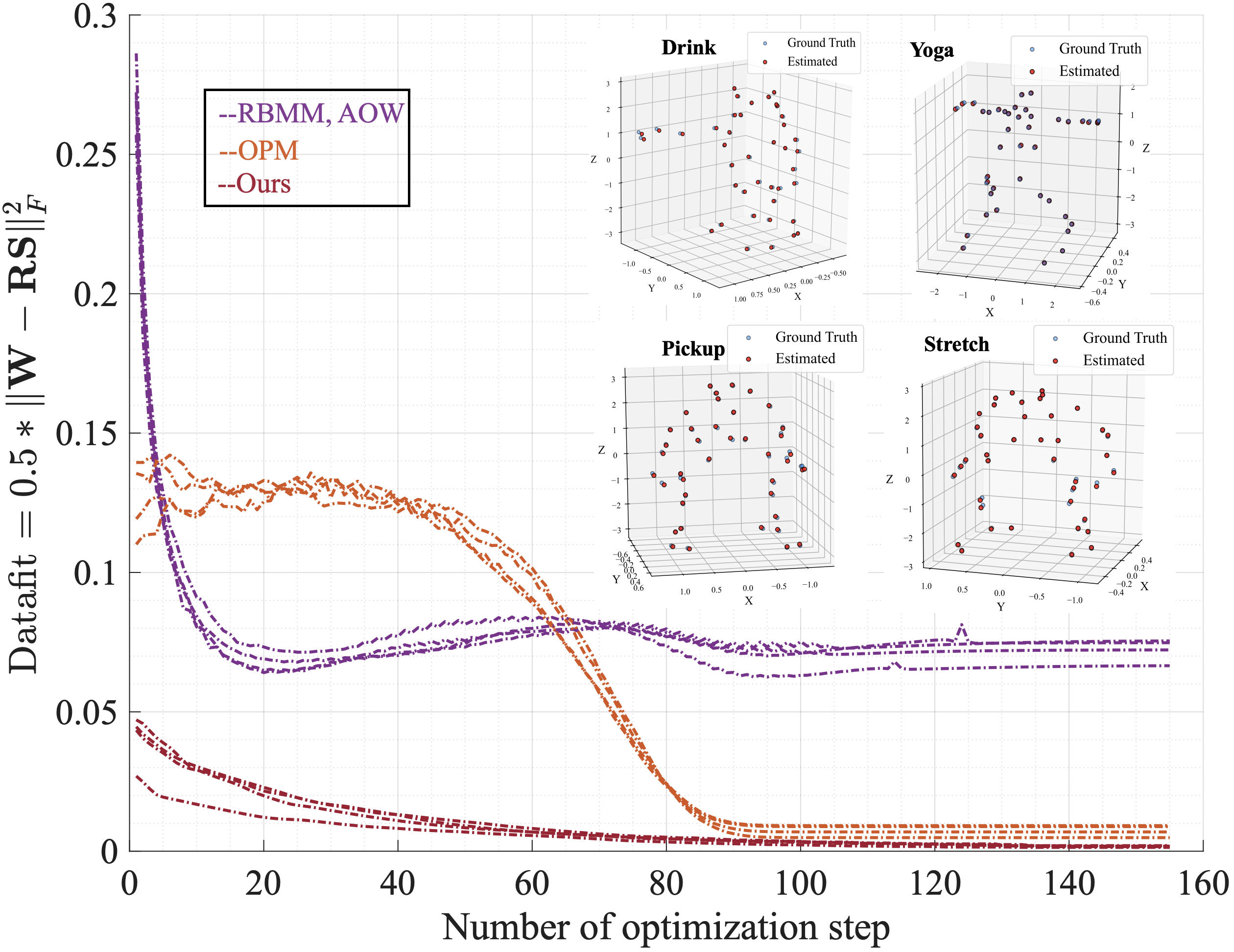}
    \caption{\scriptsize \revisedone{Stable convergence of the proposed solver on CMU MoCap sequences. The data-fit values $f(\mathbf{S})$ for R-BMM~\cite{kumar2020non}, AOW~\cite{iglesias2020accurate}, and OPM~\cite{kumar2022organic} are normalized for visualization. Qualitative 3D reconstruction results obtained using our solver are also shown for representative sequences. The 4 curves correspond to data-fit values on 4 standard CMU sequences.}}
    \label{fig:cmu_mocap_convergence}
\end{figure}

\smallskip
\noindent
\textbf{(a) Rotation Estimation.} Table~\ref{tab:rotation_error} reports our $e_{R}$ with compute time for comparison with other competing SOTA methods. Table~\ref{tab:rotation_error} shows that our method attains accuracy better or close to SOTA shape-space based methods yet demonstrating a marked advantage in computational efficiency.  For instance, on the Drink sequence, \revisedone{which has} 1102 frames and is nearly $3\times$ longer than the other sequences in Table~2, the runtime of OPM~\cite{kumar2022organic} and TSM~\cite{shi2024non} increases substantially, whereas \revisedone{our solver exhibits only a modest increase} in compute time, highlighting its scalability to longer sequences. \revisedone{For sequences without GT rotations, the supplement further reports row-orthonormality residuals
$\|\mathbf{R}_i\mathbf{R}_i^T-\mathbf{I}_2\|_F$ (refer Figure \ref{fig:rotation_constraint_plot}) and a rotation-transfer experiment in which our recovered rotations are used inside existing shape solvers (refer Table \ref{tab:rotationinit_other_method})}.

\begin{table}
\centering
\resizebox{1.0\linewidth}{!}{
\begin{tabular}{c|c|c|c|c|c}
\hline
Data & BMM \cite{dai2014simple}   & R-BMM \cite{kumar2020non} & OPM \cite{kumar2022organic} & TSM \cite{shi2024non} & Ours \\ \hline
Drink      & \makecell{0.0266 \\ (61.22)} & \makecell{0.0119 \\ (38.75)}   & \makecell{0.0071 \\ (36.08)} & \makecell{0.0033 \\ (142.96)} & \makecell{0.0166 \\ \textcolor{codebrown}{\textbf{(06.45)}}}  \\ \hline
Pickup   & \makecell{0.1731 \\ (26.95)}    &  \makecell{0.0198 \\ (11.44)}   & \makecell{0.0152 \\ (10.34)} & \makecell{0.0126 \\ (12.31)} & \makecell{0.0267  \\ \textcolor{codebrown}{\textbf{(04.88)}}}  \\ \hline
Yoga     & \makecell{0.1150 \\ (12.87)}   & \makecell{0.0129 \\ (07.60)}  & \makecell{0.0122 \\ (07.51)} & \makecell{0.0118 \\ (09.62)} & \makecell{0.0148  \\ \textcolor{codebrown}{\textbf{(04.19)}}} \\ \hline
Stretch  & \makecell{0.1034 \\ (15.28)}   &  \makecell{0.0144 \\ (09.32)}   & \makecell{0.0124 \\ (08.73)} & \makecell{0.0131 \\ (14.24)} & \makecell{0.0128  \\ \textcolor{codebrown}{\textbf{(04.21)}}} \\ \hline
Dance & \makecell{0.1864 \\ (68.68)}   &  \makecell{0.1499 \\ (12.82)}   & \makecell{0.1209 \\ (8.98)} & \makecell{0.0921 \\ (11.39)} & \makecell{0.1218 \\ \textcolor{codebrown}{\textbf{(6.39)}}} \\ \hline
Face  & \makecell{0.0303 \\ (18.68)}   &  \makecell{0.0179 \\ (7.97)}   & \makecell{0.0145 \\ (6.140)} & \makecell{0.0144 \\ (14.09)} & \makecell{0.0179 \\ \textcolor{codebrown}{\textbf{(4.66)}}} \\ \hline
Walking  & \makecell{0.1298 \\ (25.72)}   &  \makecell{0.0882 \\ (14.40)}   & \makecell{0.0816 \\ (11.26)} & \makecell{0.0710 \\ (13.64)} & \makecell{\textbf{0.0592} \\ \textcolor{codebrown}{\textbf{(4.84)}}} \\ \hline
Shark  & \makecell{0.2311 \\ (777.56)}   &  \makecell{0.0551 \\ (11.46)}   & \makecell{0.0551 \\ (7.62)} & \makecell{0.0258 \\ (13.26)} & \makecell{0.0540 \\ \textcolor{codebrown}{\textbf{(9.51)}}} \\ \hline
\end{tabular}
}
\caption{\scriptsize $e_{3d}$ and \revisedone{shape-stage runtime comparison with high-performance methods after rotations are fixed}. Best $K$ from the previous methods is used to compare $e_{3d}$ and time. Our approach is faster and its $e_{3d}$ compares favorably with competing methods. Numbers in parentheses denote computation time in seconds. }\label{tab:shape_error_1}
\end{table}

\smallskip
\noindent
\textbf{(b) Shape Estimation.}
\revisedone{
Table~\ref{tab:shape_error_1} shows the intended accuracy-runtime tradeoff. In addition to the favorable performance, our solver is consistently faster and remains competitive across the benchmark, supporting the goal of a fast no-cone, no-dual shape-space solver. For instance, on the drink sequence that has more than 1000 frames, our solver provides results in significantly less compute time.
}

Figure \ref{fig:cmu_mocap_convergence} provides qualitative results as well as convergence behaviors in comparison with recent ADMM based methods such as RBMM \cite{kumar2020non}, AOW \cite{iglesias2020accurate}, and OPM \cite{kumar2022organic}. The curves show data-fit values ($f(\mathbf{S})$) over iterations for different sequences, demonstrating stable convergence of our solver. Additionally, the plot demonstrates our approach robustness and computational efficiency relative to high-performing methods. \textit{For more results and comparison with trajectory space methods refer supplementary material.}

\smallskip
\formattedparagraph{\textit{(ii)} \nrsfm ~CVPR 2017 Benchmark.} This is a comprehensive benchmark dataset for evaluating \nrsfm~ algorithms proposed by Jensen et al. \cite{jensen2018benchmark}. It consists of 5 distinct deformable subjects, namely, Articulated, Paper, Balloon, Stretch, and Tearing. Each subject exhibits a wide range of non-rigid deformations. The deformations are recorded under 6 camera trajectories, namely, circle, flyby, line, semi-circle, tricky, and zigzag, making this dataset exceptionally challenging with a diverse set of motion and viewpoint variations. For assessment, the dataset contains ground-truth 3D shape for one reference frame only per subject category.

Table \ref{tab:nrsfm_challenge_table} presents 3D reconstruction error (root mean square error (RMSE) in millimeters \cite{taylor2010non,jensen2018benchmark}) averaged over 6 camera trajectory types. All evaluations are performed on the orthographic sequence of the dataset. The comparison includes recent as well as classical SOTA methods. \revisedone{Our solver is best or near-best on Articulated, Balloon, and Tearing, and favorably accurate on Paper and Stretch. These weaker cases involve challenging deformation and camera configurations where methods with stronger trajectory, temporal, or deformation priors can be advantageous}. The main benefit remains favorable runtime and robustness across categories (refer Figure \ref{subfig:timing_nrsfm_challenge}). Figure~3 presents qualitative results, showing the reconstructed 3D shapes alongside the ground-truth 3D shapes. \revisedone{Overall}, these results \revisedone{demonstrate the solver's speed, stability, and reconstruction quality} under complex deformations and varying camera trajectories.

\begin{figure}[t]
    \centering
    \includegraphics[width=0.9\linewidth]{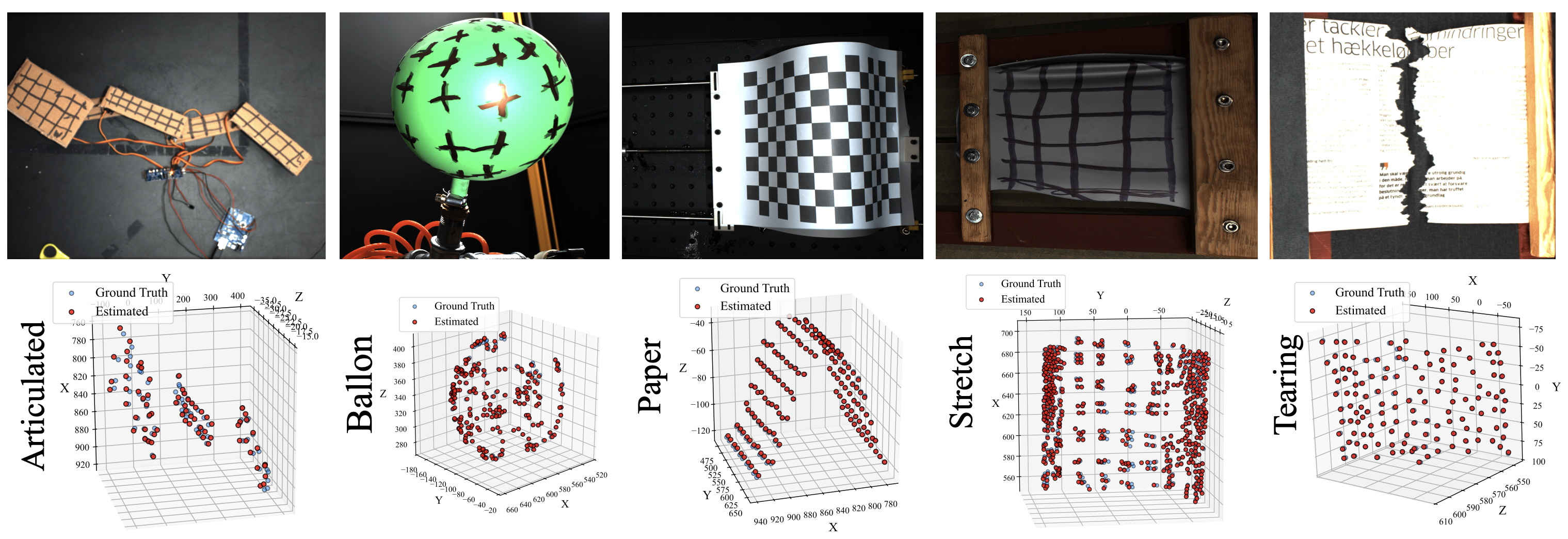}
    \caption{\scriptsize Qualitative results on \nrsfm ~challenge dataset.}
    \label{fig:nrsfm_challenge_result_visualization}
\end{figure}

\begin{table}
\centering
\resizebox{1.0\linewidth}{!}{
\begin{tabular}{c|c|c|c|c|c|c|c|c}
    \hline
    Data  & {\small CSF2~\cite{gotardo2011non}}   & {\small BMM~\cite{dai2014simple}}  &{\small R-BMM~\cite{kumar2020non}} &{\small AOW~\cite{iglesias2020accurate}}  & {\small BP~\cite{ornhag2021bilinear}}   & {\small OPM~\cite{kumar2022organic}}  & TSM \cite{shi2024non} & Ours  \\ \hline
    Articulated & \makecell{11.52} & \makecell{18.49} & \makecell{16.00} & \makecell{15.03} & \makecell{16.10} & \makecell{12.18} & \makecell{10.69} & \makecell{\textbf{10.58}}  \\ \hline %
    Balloon  & 10.14 & 10.39 & 7.84  & 8.05  & 8.29  & 6.29  & 7.28 & \makecell{\textbf{5.86} } \\ \hline %
    Paper    & \makecell{9.72} & \makecell{8.94}  & \makecell{10.69} & \makecell{10.45} & \makecell{\textbf{6.70}}  & \makecell{8.86}  & \makecell{7.91} & \makecell{10.42 }   \\ \hline %
    Stretch  & \makecell{8.65} & \makecell{10.02} & \makecell{7.53}  & \makecell{9.01}  & \makecell{7.66}  & \makecell{6.36}  & \makecell{\textbf{5.43}} & \makecell{7.02 }  \\ \hline %
    Tearing  & \makecell{12.04} & \makecell{14.23} & \makecell{16.34} & \makecell{16.20} & \makecell{11.26} & \makecell{10.91} & \makecell{10.77} & \makecell{\textbf{10.61} }  \\ \hline
    \end{tabular}
}
\caption{\scriptsize 3D reconstruction error comparison with state-of-the-art on \cite{jensen2018benchmark} dataset.}\label{tab:nrsfm_challenge_table}
\end{table}

\begin{table}
    \centering
    \resizebox{1.0\linewidth}{!}{
    \begin{tabular}{c|c|c|c|c|c|c|c|c}
        \hline
        Data  & {\small SMSR~\cite{ansari2017scalable}}   & {\small DS~\cite{dai2017dense}}  &{\small GM~\cite{kumar2018scalable}} &{\small ND~\cite{sidhu2020neural}}  & {\small DV~\cite{garg2013dense}}   & {\small OPM~\cite{kumar2022organic}}  & TSM \cite{shi2024non} & Ours  \\ \hline
        Seq.1 & \makecell{0.1893} & \makecell{0.0636} & \makecell{0.0443} & \makecell{-} & \makecell{0.0531} & \makecell{0.0624 \\ (55.60)} & \makecell{0.2223 \\ (10964.22)} & \makecell{0.0774 \\ \textcolor{codebrown}{\textbf{(8.29)}}}  \\ \hline %
        Seq.2 & \makecell{0.2133} & \makecell{0.0569} & \makecell{0.0381}  & \makecell{-}  & \makecell{0.0457}  & \makecell{0.0451 \\ (53.87)}  & \makecell{0.1271 \\ (10500.35)} & \makecell{0.0442 \\ \textcolor{codebrown}{\textbf{(8.18)}}} \\ \hline %
        Seq.3    & \makecell{0.1345} & \makecell{0.0374}  & \makecell{0.0294} & \makecell{0.0450 \\ (3h+ train)} & \makecell{0.0346}  & \makecell{0.0279 \\ (100.72)}  & \makecell{0.0393 \\ (11624.89)} & \makecell{0.0382 \\ \textcolor{codebrown}{\textbf{(15.01)}}}   \\ \hline %
        Seq.4  & \makecell{0.0984} & \makecell{0.0428} & \makecell{0.0309}  & \makecell{0.0490 \\ (3h+ train)}  & \makecell{0.0379}  & \makecell{0.0419 \\ (102.19)}  & \makecell{0.0615 \\ (11609.17)} & \makecell{0.0522 \\ \textcolor{codebrown}{\textbf{(15.16)}}}  \\ \hline %
        \end{tabular}
    }
    \caption{\scriptsize $e_{3d}$ and \revisedone{overall run-time} comparison with high-performing approaches on dense sequences \cite{garg2013dense}. Timings for other methods are unavailable due to the lack of open-source code. \textbf{Observation:} Our solver is faster than baselines: OPM and TSM.
    }\label{tab:dense_nrsfm_result}
\end{table}

\smallskip
\formattedparagraph{\textit{(iii)} Dense \nrsfm ~Benchmark.} Introduced by Garg et al. \cite{garg2013dense, garg2013variational}, this dataset comprises of four synthetic face sequences with 28,887 tracked feature points. Face Seq.1 and Face Seq.2 are 10-frame-long whereas Face Seq.3 and Face Seq.4 are 99-frame long. Table \ref{tab:dense_nrsfm_result} provides the statistical results of our approach compared to SOTA dense \nrsfm ~methods. We use $K=1$ to evaluate our approach on all the four sequences. 
Dense methods exploiting image-grid or spatial smoothness or latent space constraint can be better on some sequences \cite{kumar2018scalable, garg2013variational, sidhu2020neural}. Our solver does not use such priors and instead targets a general dense and sparse tracked-point setting with small as well as long sequences, yielding major runtime gains. \revisedone{Our computation time is \(7\times\)--\(1000\times\) faster than the reported OPM \cite{kumar2022organic} and TSM \cite{shi2024non} factorization baselines, demonstrating the scalability of the proposed solver on dense tracks}. Note: For TSM, the default initialization for dense sequence leads to code break-down so we used CSF2 \cite{gotardo2011non} initialization given in their code. \revisedone{Qualitative result on real dense sequence is provided in the supplementary}

\smallskip
\formattedparagraph{\textit{(iv)} H3WB Dataset.} This dataset \cite{zhu2023h3wb} extends Human3.6M \cite{ionescu2013human3} with 133 annotated keypoints across the full body and includes diverse daily actions. \cite{shi2024non} created a new \nrsfm ~evaluation benchmark using different camera motion types. Table~6 shows the $e_{3d}$ comparison \revisedone{on} the one-circle camera-motion setting. Our \revisedone{solver} obtains the best result on several actions and remains competitive on the rest. Additional H3WB results are provided in the supplementary material.

\begin{figure}[t]
    \centering
    \begin{subfigure}{0.23\textwidth} 
        \centering
        \includegraphics[width=\textwidth]{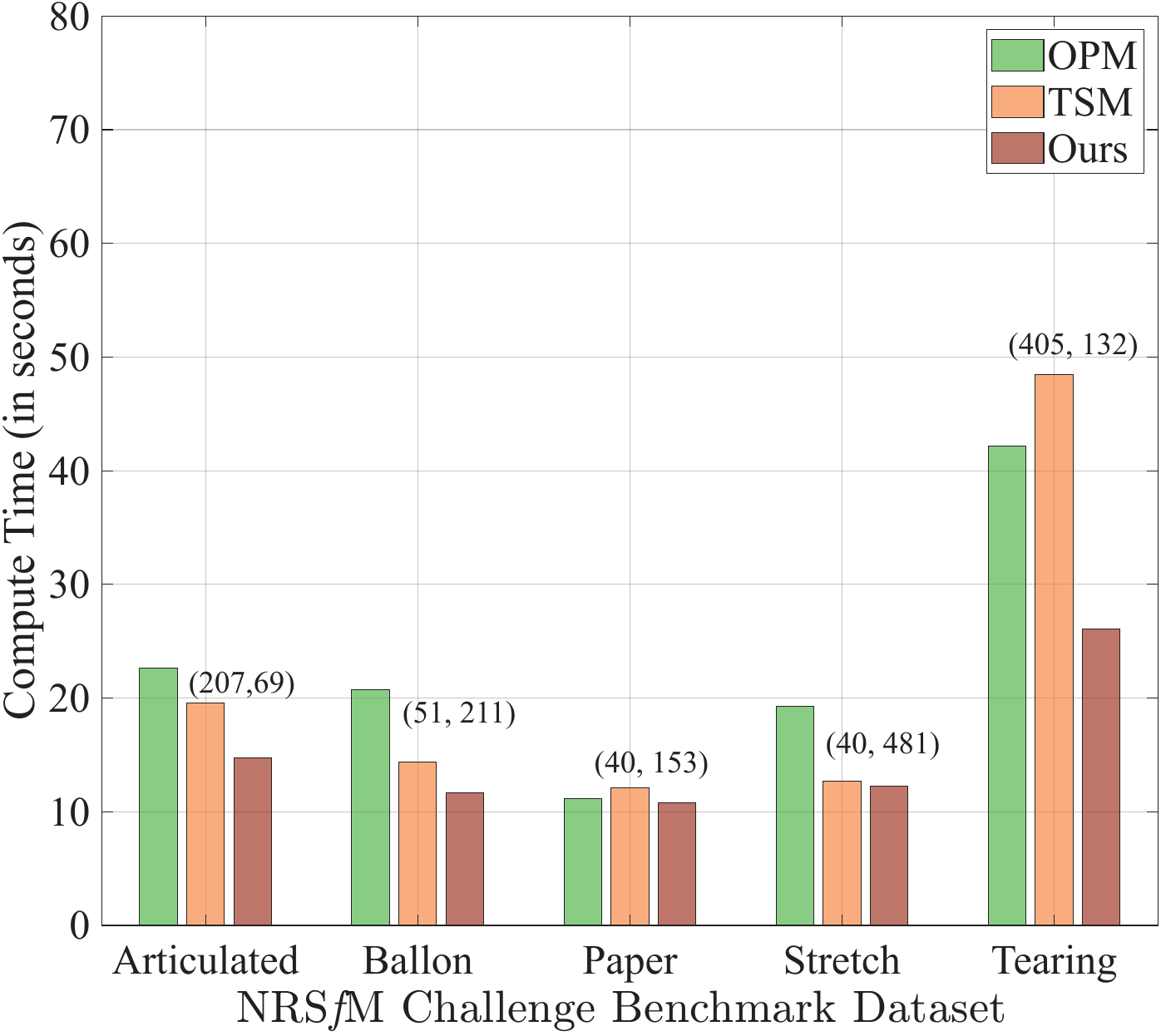} 
        \caption{Compute time comparison}\label{subfig:timing_nrsfm_challenge}
    \end{subfigure}
    \hfill 
    \begin{subfigure}{0.23\textwidth}
        \centering
        \includegraphics[width=\textwidth]{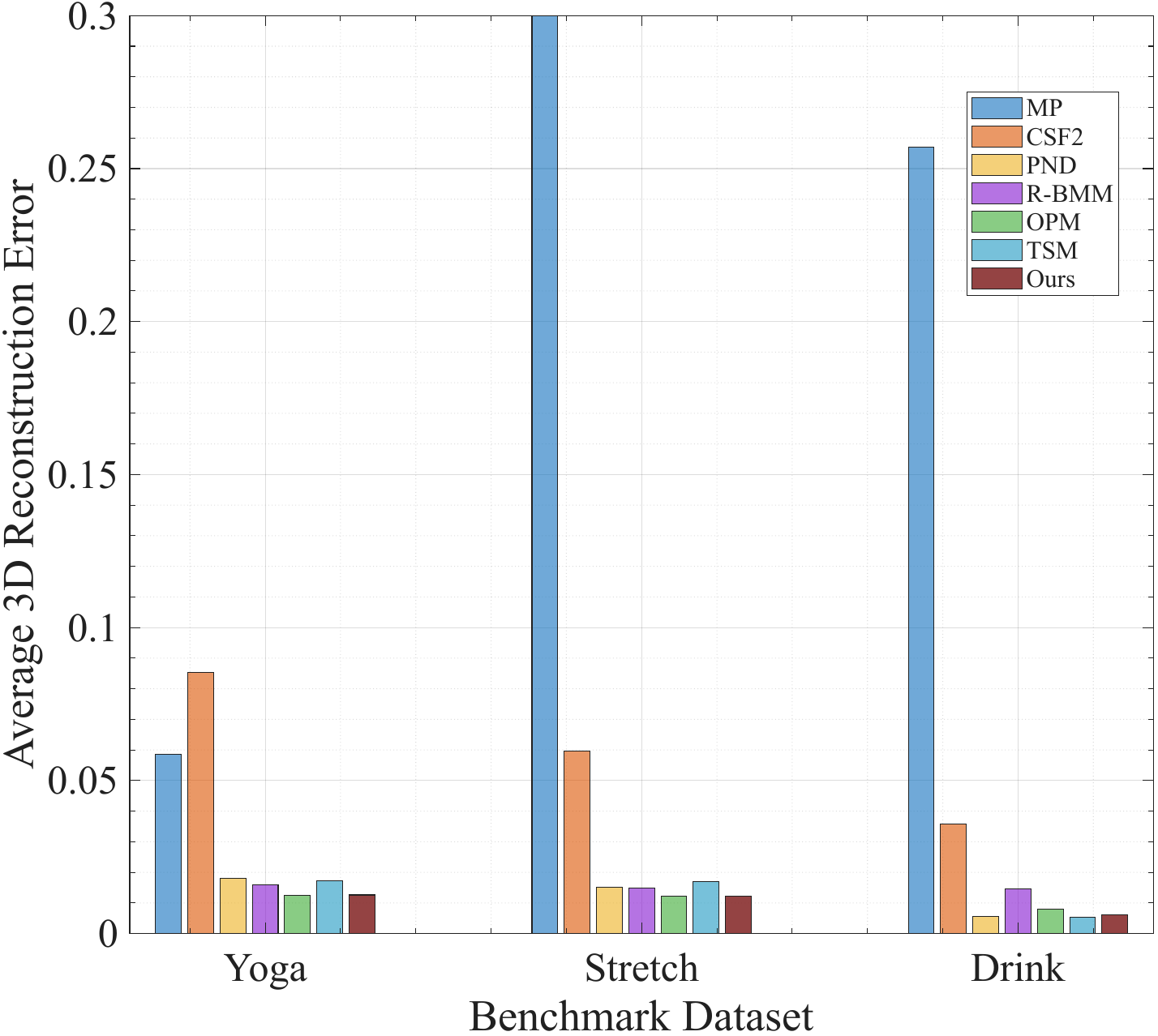}
        \caption{Results on missing data}\label{subfig:missing_data_result}
    \end{subfigure}
    \caption{\scriptsize (a) Overall compute time comparison with current SOTA approaches \cite{shi2024non, kumar2022organic} on \cite{jensen2018benchmark} dataset. \textbf{Observation}: Our approach compute time is significantly better as the  number of frames and points increases (for e.g., Tearing seq. with $F = 405$, $P = 132$). (b) Our approach compares favorably with other methods on missing dataset sequence.} 
    \label{fig:ablation1}
\end{figure}

\begin{table}[t]
\centering
\resizebox{1.0\linewidth}{!}{
\begin{tabular}{c|c|c|c|c|c|c|c}
    \hline
    Data    & {\small CSF2~\cite{gotardo2011non}} & {\small BMM~\cite{dai2014simple}} & {\small PND~\cite{lee2013procrustean}} & {\small CNS~\cite{lee2016consensus}} & {\small R-BMM~\cite{kumar2020non}} & TSM \cite{shi2024non} & Ours  \\ \hline
    Eating2 & 0.0923 & 0.0665 & 0.0701 & 0.0823 & 0.0633 & \textbf{0.0520} & 0.0651 \\ 
    Smoking1  & 0.1230 & 0.0844  & 0.1049  & 0.1073  & 0.0746  & 0.0738 & \textbf{0.0726}  \\ 
    Directions  & 0.0787  & 0.0648 & 0.0536 & {0.0621}  & 0.0595  & \textbf{0.0529} & 0.0643  \\ 
    Smoking  & 0.0702 & 0.0492  & 0.0622  & 0.0578  & 0.0363  & 0.0352 & \textbf{0.0337}  \\ 
    Waiting2  & 0.1029 & 0.1081 & 0.1114 & 0.1142 & 0.1117 & 0.1035 & \textbf{0.1012} \\ \hline
    \end{tabular}
}
\caption{\footnotesize $e_{3d}$ on One-circle-type camera motion dataset \cite{zhu2023h3wb}.
}\label{tab:table_h3wb_o}
\end{table}

\subsection{Ablations}
\begin{figure}[t]
    \centering
    \includegraphics[width=0.7\linewidth]{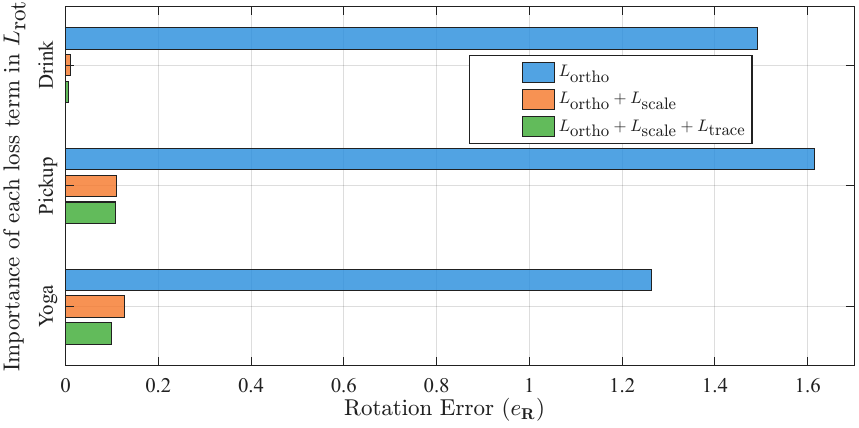}
    \caption{\scriptsize Importance of each loss term included  in Eq.\eqref{eq:overall_rotation_loss}. Variation in $e_\mathbf{R}$ as we add the three loss terms.}
    \label{fig:rotation_loss_ablation_main}
\end{figure}

\formattedparagraph{\textit{(i)} Relevance of each term in rotation loss function.}
We performed this ablation on CMU MoCap dataset \cite{akhter2009nonrigid} to demonstrate the utility of each term included in the rotation loss function---refer Eq.\eqref{eq:overall_rotation_loss}. Our rotation loss is composed of three main loss terms 1) Orthonormality loss 2) Scale loss, and 3) Trace loss. Figure \ref{fig:rotation_loss_ablation_main} shows the change in $e_\mathbf{R}$ values as we add the proposed loss terms thereby demonstrating the usefulness of each term included in the proposed loss function for rotation estimation.

\begin{figure}[t]
    \centering
    \begin{subfigure}{0.23\textwidth} 
        \centering
        \includegraphics[width=\textwidth]{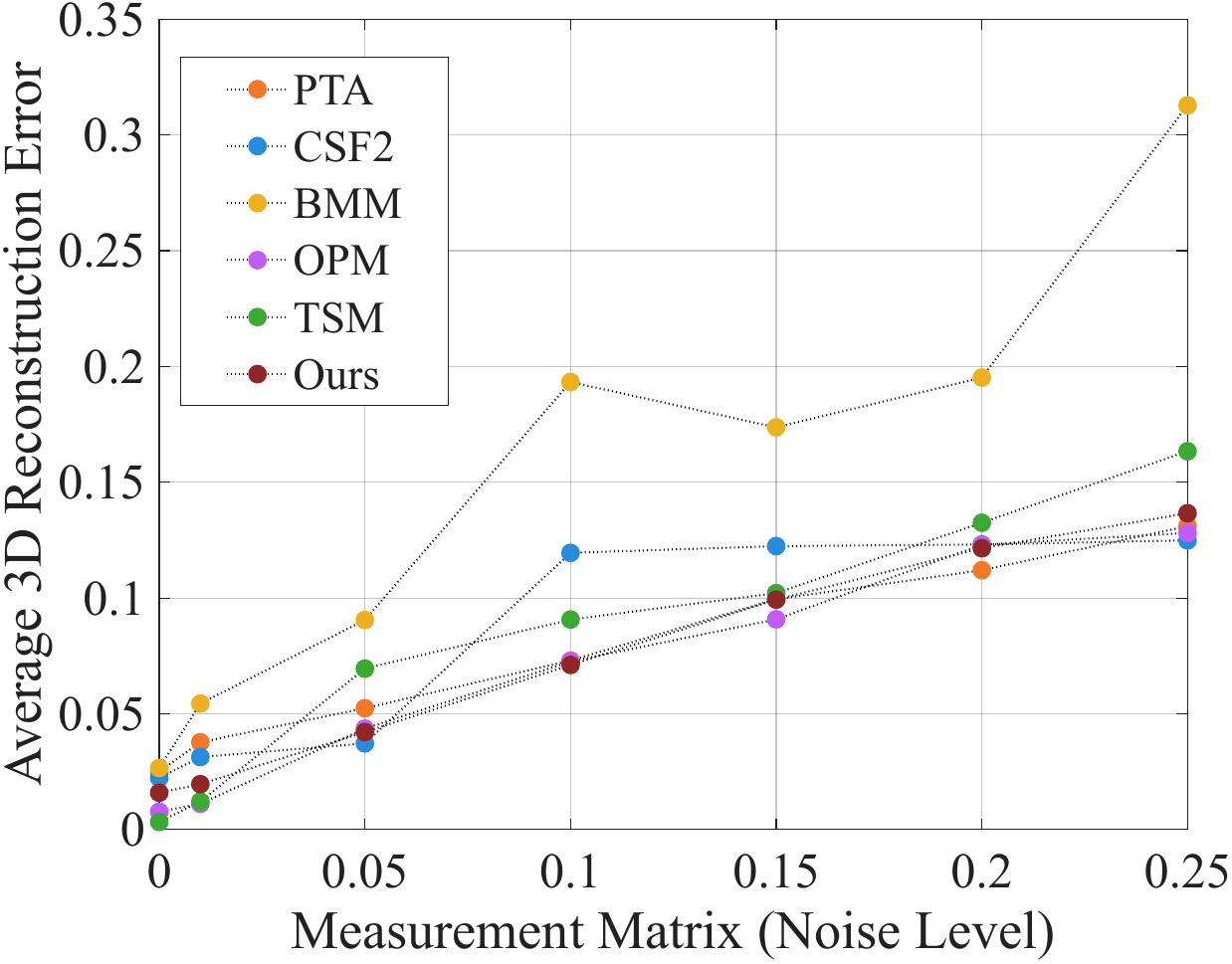} 
        \caption{$e_{3d}$ on noisy Drink seq.}\label{subfig:noisy_data_drink}
    \end{subfigure}
    \hfill 
    \begin{subfigure}{0.23\textwidth}
        \centering
        \includegraphics[width=\textwidth]{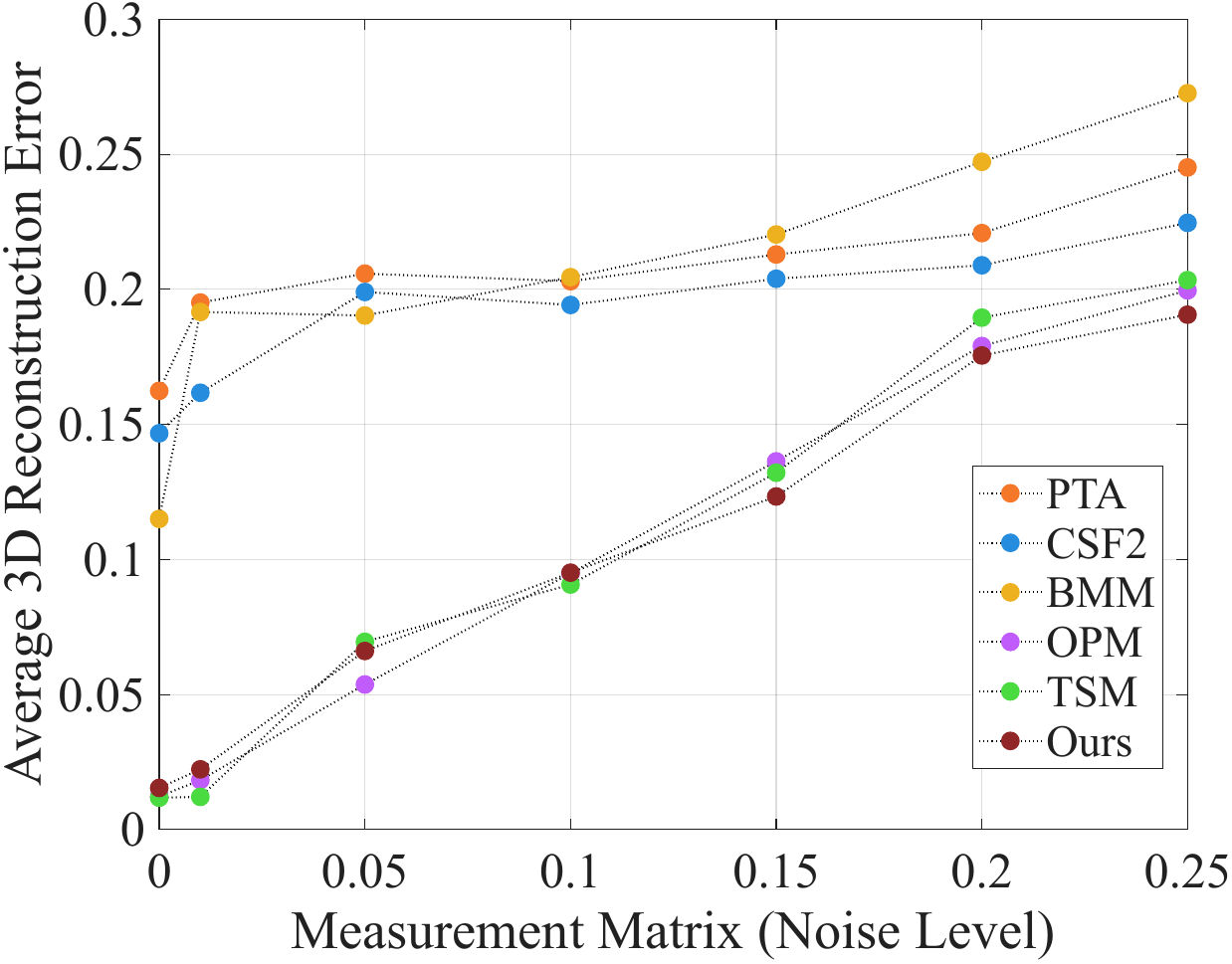}
        \caption{$e_{3d}$ on noisy Yoga seq.}\label{subfig:noisy_data_yoga}
    \end{subfigure}
    \caption{\footnotesize $e_{3d}$ variation w.r.t different noise levels in $\mathbf{W}$.} 
    \label{fig:ablation2}
\end{figure}

\smallskip
\formattedparagraph{\textit{(ii)} Performance on missing and noisy data sequence.}
For missing trajectory sequence, we used setup as detailed in \cite{kumar2022organic, cabral2013unifying}, i.e., randomly set 30\% of the trajectory missing from the $\mathbf{W}$. The results are shown in Figure \ref{subfig:missing_data_result}. Our method performs comparably with the SOTA approaches. For noisy trajectory sequence we again followed the experimental setup as described in \cite{kumar2022organic}. We executed different methods 10 times for each standard deviation value and plotted the method's mean. Figure \ref{subfig:noisy_data_drink} and Figure \ref{subfig:noisy_data_yoga} show the comparison of different methods for $e_{3d}$ on the noisy trajectory. Statistical analysis shows that our approach is robust to noise without noise-specific tuning and remains stable compared with SOTA approaches. \textit{Refer to the supplementary material for more ablation.}

\section{Limitations and Closing Remarks}
\label{sec:limitations-conclusion}

\formattedparagraph{Limitations.} The proposed solver operates in the classical orthographic, tracked-point \nrsfm ~setting. It assumes reliable 2D tracks, known visibility masks for missing data, and mean-centered measurements. Its shape stage approach does not use image-grid smoothness \cite{garg2013dense, kumar2019superpixel, liu2023vadepthnet}, explicit camera-trajectory or temporal deformation priors \cite{park20153d, kumar2019dense}, or category-level training \cite{deng2025deep, novotny2019c3dpo,  liu2023single}. This makes the solver broadly applicable, but specialized methods can be more accurate when their additional priors match the data, e.g., smooth dense surfaces or challenging camera motion. Extending our approach to perspective cameras and lightweight spatial or temporal priors is a natural future direction.

\smallskip
\formattedparagraph{Closing Remarks.} This work shows that shape-space \nrsfm~factorization need not depend on SDP cone solvers or ADMM dual variables for accurate rotation and shape recovery. It is shown that we can retain the well-founded algebraic principle of \nrsfm~factorization while making its computations simple, scalable, and accelerator-friendly by replacing Gram-matrix cone optimization with factor-side differentiable correction, and replacing ADMM with an unrolled spectral-proximal solver. More broadly, the paper points to a practical direction for geometric vision, i.e., modernize the solver while preserving the algebraic principle.
{
    \small
    \bibliographystyle{ieeenat_fullname}
    \bibliography{main}
}
\clearpage
\setcounter{page}{1}
\maketitlesupplementary

\begin{abstract}
This supplementary material provides additional experimental analysis, ablation studies, and theoretical details. Continuing from the main paper, we first present an experimental evaluation on the remaining benchmark datasets, including a statistical comparison with trajectory-space methods, which at present are not SOTA in terms of 3D reconstruction accuracy but are useful approaches. Next, we compare our approach's performance with recent deep learning methods on the CMU long-sequence dataset. Furthermore, the supplement presents essential results from the study, including what happens when we feed our rotation estimates to current state-of-the-art methods, as well as additional experimental observations on noisy data sequences and overall solver pipeline. On the theoretical side, we first provide a step-by-step derivation for Eq.\eqref{eq:per_frame_rot} from the main paper. Next, extensive details regarding the complexity of the proposed \nrsfm ~solver compared to the recent SOTA methods are presented. At the end, we prove the Lipschitz continuity of the reprojection error term in Eq.\eqref{eq:shape_optimization} and present additional qualitative results for completeness.
\end{abstract}

\section{Additional Results on Benchmark Dataset}

\subsection{Comparison with trajectory-space approaches}
\revisedone{Trajectory-space methods are important efficient alternatives to shape-space factorization. They are often fast because they reduce the number of unknowns through fixed temporal bases, trajectory priors, or probabilistic trajectory distributions, rather than solving for a full shape-space representation. Therefore, we compare them primarily in terms of reconstruction accuracy and representation assumptions. Our main runtime claim is restricted to recent shape-space SDP/ADMM baselines, where the proposed no-cone no-dual solver removes the dominant optimization bottlenecks.}

\smallskip
\formattedparagraph{\textit{(i)} On CMU MoCap Dataset.} Trajectory space methods uses known DCT-basis to solve \nrsfm ~\cite{akhter2009nonrigid, gotardo2011computing}, thereby significantly reducing the number of unknowns to be solved. Therefore, by design these methods are generally faster in compute time than shape space methods, yet fails to model high-frequency details. Table \ref{tab:cmu_mocap_table_trajectoryspace} provides the $e_{3d}$ statistical comparison with popular trajectory space approaches on CMU MoCap \cite{akhter2009nonrigid} and Torresani et al. \cite{torresani2004learning} dataset. Based on the statistical results, it is easy to observe that more often than not our approach performs better.

\begin{table}[h]
\centering
\resizebox{1.0\linewidth}{!}{
\begin{tabular}{c|c|c|c|c|c|c|c|c}
\hline
Data    & CSF1~\cite{gotardo2011computing}  & CSF2~\cite{gotardo2011non}  & KSTA~\cite{gotardo2011kernel}  & PTA~\cite{akhter2009nonrigid} & NP~\cite{wang2022neural} & PBezier \cite{agudo2021piecewise} &  PPTA \cite{agudo2018scalable} &  Ours     \\ \hline
Drink   & 0.0223 & 0.0223 & {0.0156} & 0.0250 & 0.0354 & \textbf{0.009} & 0.011 & 0.0166 \\
Pickup  & 0.2301 & 0.2277 & 0.2322 & 0.2369 & 0.0284 & 0.138 & 0.235  & \textbf{0.0267} \\
Yoga    & 0.1467 & 0.1464 & 0.1476 & 0.1624 & 0.0210 & 0.111 & 0.158 & \textbf{0.0148} \\
Stretch & 0.0710 & 0.0685 & 0.0674 & 0.1088 & 0.0151 & 0.062 &  0.084 & \textbf{0.0128} \\
Dance   & 0.2705 & 0.1983 & 0.2504 & 0.2960 & \textbf{0.0980} & 0.143 & 0.229 & 0.1218 \\
Face    & 0.0363 & 0.0314 & 0.0339 & 0.0436 & 0.0352 & -- & -- & \textbf{0.0179} \\
Walking & 0.1893 & 0.1035 & 0.1029 & 0.3951 & 0.1223 & -- & -- & \textbf{0.0592} \\
Shark  & \textbf{0.0081} & 0.0444 & 0.0160 & 0.1804 & 0.0942 & -- & -- & 0.0540 \\ \hline
\end{tabular}
}
\caption{
$e_{3d}$ comparison on MoCap datasets~\cite{akhter2009nonrigid,torresani2004learning} with trajectory-space and trajectory-prior NRSfM methods. 
Piecewise B\'ezier~\cite{agudo2021piecewise} and PPTA~\cite{agudo2018scalable} are included as efficient trajectory-prior alternatives. Best performance for each sequence is highlighted in bold. 
\revisedone{PBezier and PPTA results are reported from their published papers because public code is unavailable; ``--'' denotes unavailable results.}
}
\label{tab:cmu_mocap_table_trajectoryspace}
\end{table}

\smallskip
\formattedparagraph{\textit{(ii)} Dense Face Dataset.} This dataset is composed of four face sequences with complex facial deformations. Table \ref{tab:dense_sequence_table_trajectoryspace} provides the $e_{3d}$ statistical comparison with popular trajectory space approaches. As stated, and can be observed from the statistical results that trajectory based approaches encounter challenges in modeling high-frequency details, ~whereas our approach performs well on all these sequences.

\begin{table}[h]
\centering
\resizebox{1.0\linewidth}{!}{
\begin{tabular}{c|c|c|c|c|c}
\hline
Data    & CSF1~\cite{gotardo2011computing}  & CSF2~\cite{gotardo2011non}  & PTA~\cite{akhter2009nonrigid}  & NP~\cite{wang2022neural} &  Ours     \\ \hline
Face Seq.1   & 0.5325 & 0.4677 & 0.1559 & - & \textbf{0.0774} \\
Face Seq.2   & 0.9266 & 0.7909 & 0.1503 & - & \textbf{0.0442} \\
Face Seq.3   & 0.5274 & 0.5474 & 0.1252 & 0.0394 & \textbf{0.0382} \\
Face Seq.4   & 0.5392 & 0.5292 & 0.1348 & 0.0574 & \textbf{0.0522} \\ \hline
\end{tabular}
}
\caption{$e_{3d}$ comparison on Dense Face Dataset with popular \textbf{trajectory space} \cite{garg2013dense}. Best performance for each sequence is highlighted in black.}\label{tab:dense_sequence_table_trajectoryspace}
\end{table}

\subsection{More results on H3WB dataset} Continuing our evaluation on H3WB dataset \cite{zhu2023h3wb} from the main paper, we further evaluated our approach on the other two camera motion provided with this dataset, i.e., Fixed-type camera motion and Multi-circle-type camera motion. Table \ref{tab:table_h3wb_w} and Table \ref{tab:table_h3wb_c} show $e_{3d}$ comparison of our approach with the SOTA \nrsfm ~matrix factorization approaches. In most of the cases for Fixed-Type camera motion, our approach achieves better $e_{3d}$ while for Multi-circle-type camera motion our approach accuracy is favorably close of the SOTA method accuracy.

\begin{table}[h]
\centering
\resizebox{1.0\linewidth}{!}{
\begin{tabular}{c|c|c|c|c|c|c|c}
    \hline
    \multicolumn{7}{c}{\textbf{Fixed-type camera motion}} \\ \hline
    Data    & {\small CSF2~\cite{gotardo2011non}} & {\small BMM~\cite{dai2014simple}} & {\small PND~\cite{lee2013procrustean}} & {\small CNS~\cite{lee2016consensus}} & {\small R-BMM~\cite{kumar2020non}} & TSM \cite{shi2024non} & Ours  \\ \hline
    Eating2 & 0.2662 & 0.2084 & 0.2309 & 0.2250 & 0.1779 & \underline{0.1767} & \textbf{0.1671} \\ %
    Smoking1  & 0.3534 & 0.2287  & 0.3397  & 0.2982  & 0.2290  & \textbf{0.2040} & \underline{0.2181}  \\ %
    Directions  & 0.2953  & 0.2872 & 0.3002 & \underline{0.2674}  & 0.2964  & 0.2688 & \textbf{0.2149}  \\ %
    Smoking  & 0.1980 & 0.1759  & 0.4349  & 0.1968  & 0.1760  & \underline{0.1693} & \textbf{0.1664}  \\ %
    Waiting2  & 0.2125 & 0.1712 & 0.3306 & 0.1848 & 0.1478 & \textbf{0.1097} & \underline{0.1161}\\ \hline
    \end{tabular}
}
\caption{$e_{3d}$ comparison on H3WB \cite{zhu2023h3wb} with Fixed-type camera motion. $2^\textrm{nd}$ best result is underlined.}\label{tab:table_h3wb_w}
\end{table}

\begin{table}[h]
\centering
\resizebox{1.0\linewidth}{!}{
\begin{tabular}{c|c|c|c|c|c|c|c}
    \hline
    \multicolumn{7}{c}{\textbf{Multi-circle-type camera motion}} \\ \hline
    Data    & {CSF2~\cite{gotardo2011non}} & {BMM~\cite{dai2014simple}} & {PND~\cite{lee2013procrustean}} & {CNS~\cite{lee2016consensus}} & { R-BMM~\cite{kumar2020non}} & TSM \cite{shi2024non} & Ours  \\ \hline
    Eating2 & 0.0658 & 0.0384 & 0.0395 & 0.0605 & 0.0335 & \textbf{0.0276} & \underline{0.0333} \\ 
    Smoking1  & 0.0507 & 0.0486  & 0.0358  & 0.0662  & 0.0332  & \textbf{0.0294} & \underline{0.0325}  \\ 
    Directions  & 0.0456  & 0.0569 & \underline{0.0322} & {0.0535}  & 0.0518  & \textbf{0.0315} & 0.0438  \\ 
    Smoking  & 0.0708 & 0.0308  & 0.0292  & 0.0346  & 0.0328  & 0.0282 & \textbf{0.0273}  \\ 
    Waiting2  &  0.0892 & 0.0742  & 0.0450  &  0.0873  & 0.0637  &  0.0390 & \underline{0.0424}  \\ 
    \hline
    \end{tabular}
}
\caption{$e_{3d}$ comparison on H3WB \cite{zhu2023h3wb} with Multi-circle-type camera motion. $2^\textrm{nd}$ best result is underlined.}\label{tab:table_h3wb_c}
\end{table}

\begin{figure*}[t]
\centering
\begin{subfigure}{0.48\textwidth}
\centering
    \includegraphics[width=\textwidth]{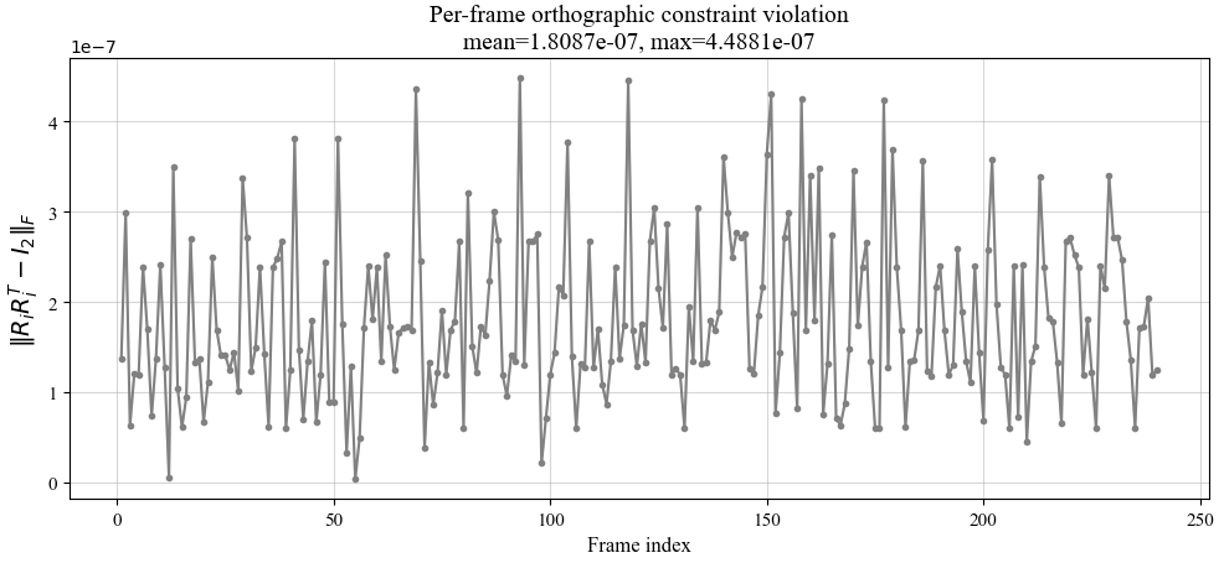}
    \caption{Shark}\label{fig:shark2-r}   
\end{subfigure}
\begin{subfigure}{0.48\textwidth}
\centering
    \includegraphics[width=\textwidth]{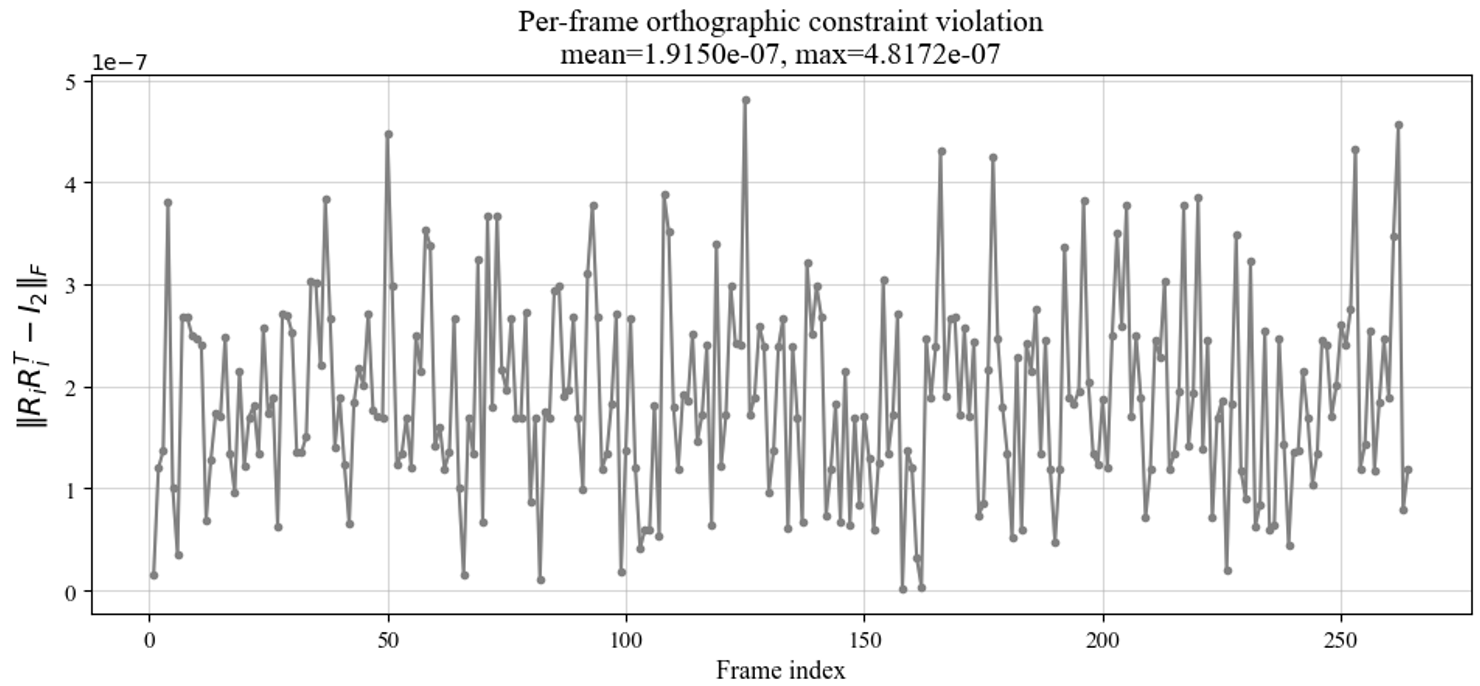}
    \caption{Dance}\label{fig:dance2-r}   
\end{subfigure}
\caption{Constraint-satisfaction analysis on CMU MoCap sequences without ground-truth rotations. For each frame, we report the row-orthonormality residual $\|\mathbf{R}_i\mathbf{R}_i^{T}-\mathbf{I}_2\|_F$ of the recovered $2\times3$ camera block after row-polar projection using Newton-Schulz inverse-square-root iterations. The plots show frame-wise residuals together with the sequence-level mean and maximum values. 
Across both the sequence, the residuals remain on the order of $10^{-7},$ confirming that the recovered camera blocks satisfy the orthographic Stiefel constraint to high numerical precision.}
\label{fig:rotation_constraint_plot}
\end{figure*}

\begin{table*}[t]
\centering
\resizebox{1.0\linewidth}{!}{
\begin{tabular}[t]{l|c|c|c|c|c|c|c|c|c}
\noalign{\hrule height 1.0pt}
Methods  & Subj. 07  & Subj. 20 & Subj. 23  & Subj. 33 & Subj. 34 & Subj. 38 & Subj. 39 & Subj. 43 & Subj. 93\\
\hline
C3DPO \cite{novotny2019c3dpo}    & 0.286 & 0.361 & 0.413 & 0.421 & 0.401 & 0.263 & 0.330 & 0.491 & 0.325 \\
DNRSFM \cite{kong2019deep}   & 0.097 & 0.219 & 0.264 & 0.219 & 0.209 & 0.137 & 0.127 &  0.223 &  0.164 \\
PR-PRN \cite{zeng2021pr}  & \textbf{0.061} &   0.167 &  0.249 & 0.254 &  0.265 & 0.108 &  \textbf{0.028} &  \textbf{0.080}  & 0.242 \\
MHR-Net \cite{zeng2022mhr} & 0.398 &   0.467 & 0.511  & 0.474 & 0.471 & 0.435 & 0.428  &  0.506 &   0.476 \\
DNRSFM-R \cite{deng2025deep} & 0.069 & 0.188 & 0.288 & 0.251 & 0.213 & \textbf{0.096} & 0.084 & 0.121 & 0.120 \\
Ours & \textbf{0.061} & \textbf{0.141} & \textbf{0.194} & \textbf{0.196} & \textbf{0.192} & \textcolor{codebrown}{\textbf{0.099}} & \textcolor{codebrown}{\textbf{0.072}} & \textcolor{codebrown}{\textbf{0.092}} & \textbf{0.097} \\
\hline
\end{tabular}
}
\caption{\small ${e_{3d}}$ on large CMU MOCAP ({UNSEEN categories}). Brown shows $2^\textrm{nd}$ best.}
\label{tab:cmu_mocap_large_scale}
\end{table*}

\subsection{Orthonormality residuals}
For sequences without ground-truth camera rotations, direct rotation-error evaluation is not possible. We therefore assess whether the recovered camera blocks satisfy the defining orthographic constraint. Specifically, for each frame, we compute
$\|\mathbf{R}_i\mathbf{R}_i^{T}-\mathbf{I}_2\|_F,$
where $\mathbf{R}_i\in\mathbb{R}^{2\times3}$ is the estimated camera block. As shown in Figure \ref{fig:rotation_constraint_plot}, on the Shark and Dance sequences, the residuals remain on the order of $10^{-7}$ across frames with similarly small mean and maximum errors. A similar trend is being observed on other dataset sequences as well. This confirms that the row-polar projection, implemented via Newton-Schulz inverse-square-root iterations, produces numerically stable row-orthonormal camera matrices even when ground-truth rotations are unavailable.

\section{Comparison with deep-learning approaches on long-sequence CMU MoCap Dataset.} This large-scale dataset contains sequences from the CMU Motion Capture database, comprising $144$ subjects, $31$ keypoints, and more than $10,000$ frames. The train and test sets are divided as detailed in \cite{kong2019deep}. Since our approach does not rely on category-specific training, we compared it with existing methods on the test set  (UNSEEN). Table \ref{tab:cmu_mocap_large_scale} shows that our \nrsfm ~solver despite learning any dataset driven object prior works well for large sequences, thus stretching the scope of factorization approach, which previously was restricted smaller sequences.

\section{Additional Ablation Study}

\formattedparagraph{\textit{(i)} Our solver rotation estimates to other methods.} 
We tested the suitability of the rotation matrix $\mathbf{R} \in \mathbb{R}^{2F \times 3F}$ estimated using our approach when integrated with the existing SOTA approaches to recover 3D shape. Table \ref{tab:rotationinit_other_method} shows the $e_{3d}$ achieved using it. Convincingly, the rotation recovered using our proposed solver help improve other method's 3D reconstruction result in most cases. For other cases, the difference in $e_{3d}$ is in the order of $10^{-3}$, thus insignificant, yet the compute time improvement can be significant \textit{c.f.}, Table \ref{tab:rotation_error}.

\begin{table}
\centering
\scriptsize
\begin{tabular}{c|c|c|c|c|c}
\hline
Data & K-value & BMM \cite{dai2014simple} & R-BMM \cite{kumar2020non} & OPM \cite{kumar2022organic} & TSM \cite{shi2024non} \\ \hline
Drink     & 11 & \makecell{\textbf{0.0164} \\ \hline 0.0266} & \makecell{\textbf{0.0084} \\ \hline 0.0119} & \makecell{0.0083 \\ \hline 0.0071} & \makecell{0.0041 \\ \hline 0.0033}  \\ \hline
Pickup    & 12 & \makecell{\textbf{0.0321} \\ \hline 0.1731} & \makecell{\textbf{0.0164} \\ \hline 0.0198} & \makecell{0.0155 \\ \hline 0.0152} & \makecell{0.0128 \\ \hline 0.0126}   \\ \hline
Yoga      & 11 & \makecell{ \textbf{0.0284} \\ \hline 0.1150} & \makecell{\textbf{0.0122} \\ \hline 0.0129} & \makecell{0.0140 \\ \hline 0.0122} & \makecell{\textbf{0.0114} \\ \hline 0.0118}  \\ \hline
Stretch   & 13 & \makecell{\textbf{0.0294} \\ \hline 0.1034} & \makecell{\textbf{0.0126} \\ \hline 0.0144} & \makecell{\textbf{0.0116} \\ \hline 0.0124} & \makecell{0.0139 \\ \hline 0.0131}  \\ \hline
\end{tabular}
\caption{Suitability of our rotation solver when used in other high-performance methods. Bold indicate improved $e_{3d}$ using our rotation estimates. For each method, \textbf{Top row} shows result using our rotation and \textbf{Bottom row} shows results reported by these methods using their rotation estimates.}\label{tab:rotationinit_other_method}
\end{table}

\smallskip
\formattedparagraph{\textit{(ii)} Additional results on noisy data sequence.} Figure \ref{fig:ablation2} shows the performance of our approach on the Pickup and Stretch sequence compared to the other SOTA methods. Our approach shows a similar trend in the $e_{3d}$ variations with different noise levels in $\mathbf{W}$. Such observations help us infer that our approach is robust in dealing with noise and well-suited for practical applications in soft robotics and intricate vision-based automation.

\begin{figure}[t]
    \centering
    \begin{subfigure}{0.23\textwidth} 
        \centering
        \includegraphics[width=\textwidth]{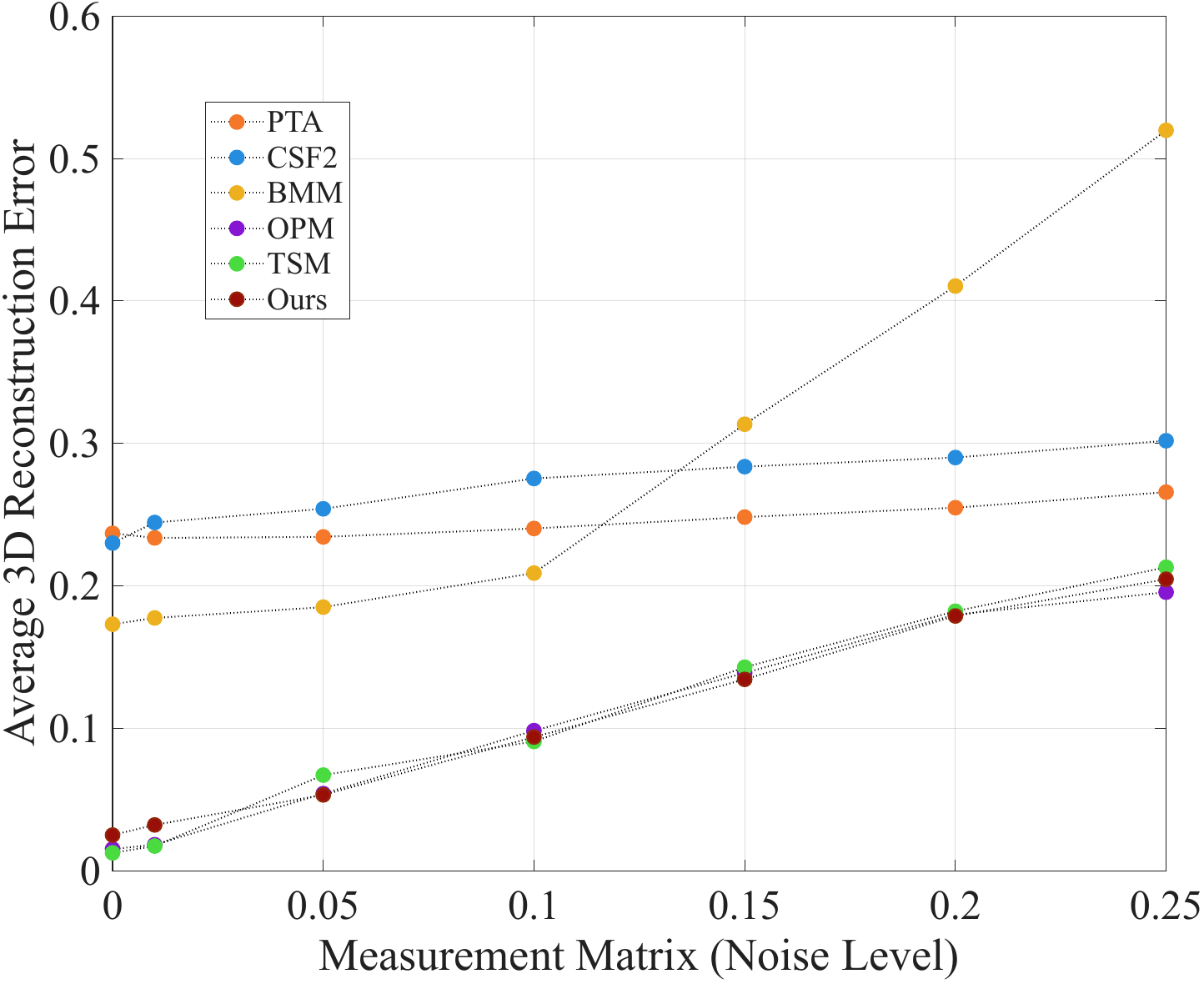} 
        \caption{$e_{3d}$ on noisy Pickup Sequence}\label{subfig:noisy_data_pickup1}
    \end{subfigure}
    \hfill 
    \begin{subfigure}{0.23\textwidth}
        \centering
        \includegraphics[width=\textwidth]{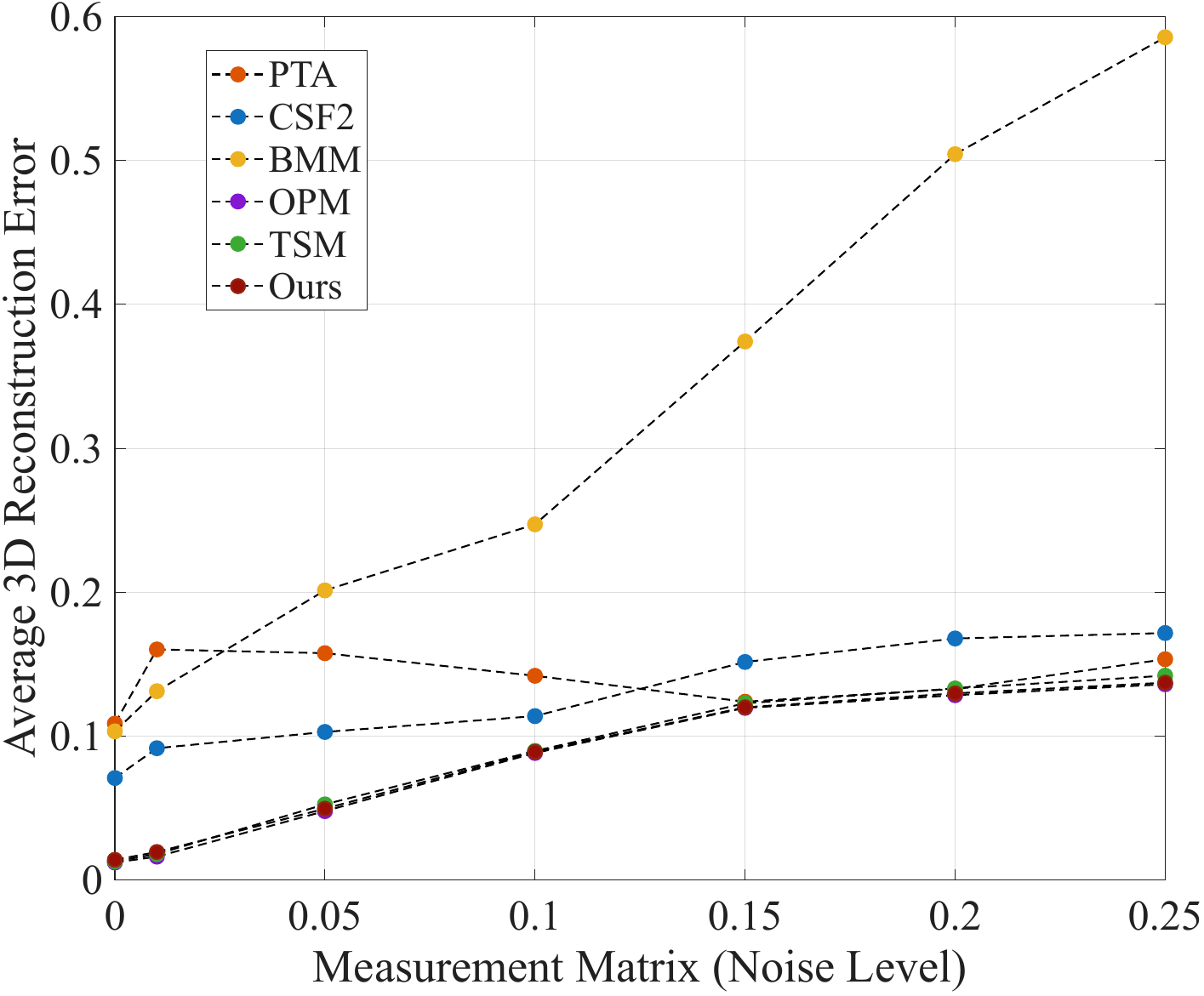}
        \caption{$e_{3d}$ on noisy Stretch Sequence}\label{subfig:noisy_data_stretch}
    \end{subfigure}
    \caption{\footnotesize $e_{3d}$ variation w.r.t different noise levels in $\mathbf{W}$.} 
    \label{fig:ablation2}
\end{figure}

\section{Detailed Pipeline}\label{sec:supp_pipeline}
\begin{figure}[t]
    \centering
    \includegraphics[width=\linewidth]{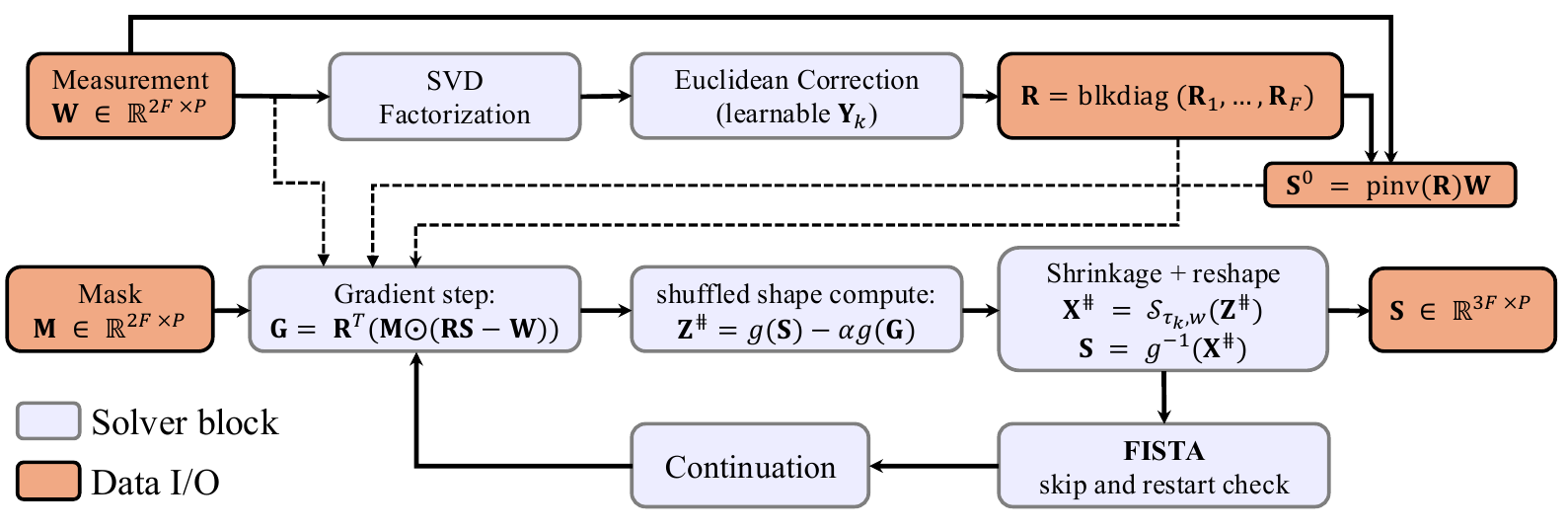}
    \caption{Detailed computational pipeline of the proposed no-cone, no-dual approach. The rotation stage performs factor-side Euclidean correction using learnable per-sequence factors $\mathbf{Y}_k$ and row-polar projection to recover valid camera blocks.  The shape stage applies a fixed unrolled spectral-proximal loop consisting of reprojection-gradient updates, reshaping to $\mathbf{S}^{\sharp}$, weighted singular-value shrinkage, FISTA-style restart, and continuation. The main computations are matrix multiplications, SVD, shrinkage, and small row-polar projections.}\label{fig:supp_approach_pipeline}
\end{figure}

Figure~\ref{fig:supp_approach_pipeline} provides a detailed computational view of the proposed solver. Starting from the tracked measurement matrix $\mathbf{W}$, we first compute a low-rank SVD factorization and estimate the block-orthographic camera matrix $\mathbf{R}$ using the factor-side Euclidean correction described in Sec.~\ref{ssec:rotation-estimation} of the main paper. Given $\mathbf{R}$, the shape stage initializes $\mathbf{S}^{0}=\mathrm{pinv}(\mathbf{R})\mathbf{W}$ and then applies the unrolled no-dual spectral-proximal solver. Each layer consists of a reprojection-gradient step, reshaping to the shape-space matrix $\mathbf{S}^{\sharp}$, weighted singular-value shrinkage, inverse reshaping, FISTA-style restart, and continuation. Thus, the full pipeline replaces SDP cone solvers for rotation and ADMM dual-variable updates for shape with a sequence of matrix multiplications, SVD-based shrinkage, and small row-polar projections.

\section{Why the row–polar projection is the closest $2 \times 3$ row–orthonormal matrix?}\label{sec:rowpolar}
Let's rewrite the original objective: 
\begin{equation}
\begin{aligned}
\arg \min_{k} \frac{1}{F} \sum_{i=1}^{F}\Big(\big(\| \mathbf{g}_{1, ik}\|_{2}^2 - \| \mathbf{g}_{2, ik}\|_{2}^2\big)^2 + \big(\mathbf{g}_{1, ik} \cdot \mathbf{g}_{2, ik}\big)^2 \Big) 
\end{aligned}\label{eq:rotation_loss_ortho_supp}
\end{equation}
In essence, this equation is trying to measure how close the two vectors $\mathbf{g}_{1, ik}$ and $\mathbf{g}_{2, ik}$ are to being orthonormal. So, Eq.\eqref{eq:rotation_loss_ortho_supp} selects the best index $k^*$ such that the corresponding $\mathbf{\Omega}_{ik^{*}} \triangleq [\mathbf{g}_{1, ik^*}; \mathbf{g}_{2, ik^*}]$ is most orthonormal. Once we selected the best one, we want to make it strictly orthonormal. That's where the notion of exact projection is invoked, which is derived next.

Given a $2 \times 3$ matrix $ \mathbf{A} = \mathbf{\Omega}_{ik^{*}}$ of row rank $ r \le 2$, we want the closest matrix $ \mathbf{\Omega}$ whose rows are orthonormal, i.e., $ \mathbf{\Omega} \in \mathrm{St}(2,3) =\{\mathbf{\Omega} \in \mathbb{R}^{2\times 3} : \mathbf{\Omega}\mathbf{\Omega}^T = \mathbf{I}_2\}$ and
\begin{equation}
    \mathbf{\Omega}^\star \in  \min_{\mathbf{\Omega} \in \mathrm{St}(2,3)} \|\mathbf{A} -  \mathbf{\Omega}\|_F^2.
\end{equation} \label{eq:Omega_main_objective}
Here, we show that the unique minimizer for full row rank is the row–polar factor

\begin{equation}
\label{eq:rowpolar}
\mathbf{\Omega}^\star =  (\mathbf{A}\mathbf{A}^T)^{-\frac{1}{2}} \mathbf{A}
\end{equation}
which is exactly the expression used in the main text for $\mathbf{R}_i$:
\begin{equation}
    \mathbf{R}_i =  \big(\mathbf{\Omega}_{ik^\star} \mathbf{\Omega}_{ik^\star}^T \big)^{-\frac{1}{2}}\mathbf{\Omega}_{ik^\star}.
\end{equation}
We also show that $\mathbf{\Omega}^\star$ is a partial isometry, i.e., row–orthonormal, and that it is the closest point in Frobenius norm on the Stiefel manifold.

\subsection{Equivalent trace maximization}
Expand the Frobenius objective Eq.\eqref{eq:Omega_main_objective}:
\[
\|\mathbf{A}-\mathbf{\Omega}\|_F^2 = \|\mathbf{A}\|_F^2 + \|\mathbf{\Omega}\|_F^2 - 2\mathbf{Tr}(\mathbf{\Omega} \mathbf{A}^T).
\]
For any $\mathbf{\Omega} \in \mathrm{St}(2,3)$, $\|\mathbf{\Omega}\|_F^2 = \mathbf{Tr}(\mathbf{\Omega}\mathbf{\Omega}^T) = \mathrm{Tr}(\mathbf{I}_2) = 2$, therefore,
\[
\|\mathbf{A} - \mathbf{\Omega}\|_F^2 = \|\mathbf{A}\|_F^2 + 2 - 2\mathbf{Tr}(\mathbf{\Omega} \mathbf{A}^T).
\]
Minimizing $\|\mathbf{A} - \mathbf{\Omega}\|_F^2$ over $\mathbf{\Omega} \in \mathrm{St}(2,3)$ is therefore {equivalent} to maximizing the linear functional
\begin{equation}\label{eq:trace-max}
\mathbf{\Omega}^\star \in \arg\max_{\mathbf{\Omega} \in \mathrm{St}(2,3)} \mathrm{Tr}(\mathbf{\Omega} \mathbf{A}^T).
\end{equation}

\subsection{SVD reduction and solution form}
Let the SVD of $\mathbf{A}$ be
\begin{equation}
    \begin{aligned}
    & \displaystyle \mathbf{A} = \mathbf{U} \mathbf{\Sigma} \mathbf{V}^T, ~\textrm{where,}  ~\mathbf{U} \in \mathbb{R}^{2\times 2}, ~ \mathbf{V} \in\mathbb{R}^{3\times 3}, ~\textrm{and}, \\
    & \displaystyle \mathbf{\Sigma} =  \begin{bmatrix}\sigma_1 & 0 & 0\\[2pt] 0 & \sigma_2 & 0 \end{bmatrix} \in \mathbb{R}^{2 \times 3},  ~~\sigma_1\ge \sigma_2\ge 0,
    \end{aligned}
\end{equation}
so that $\mathbf{\Sigma}$ has possibly zero third column. Let's write
\[
\mathbf{Tr}(\mathbf{\Omega} \mathbf{A}^T) =  \mathbf{Tr}(\mathbf{\Omega} \mathbf{V} \mathbf{\Sigma}^T \mathbf{U}^T) = \mathbf{Tr}( \mathbf{U}^T \mathbf{\Omega} \mathbf{V}\mathbf{\Sigma}^T).
\]
By defining the change of variables
\[
\mathbf{B} \triangleq \mathbf{U}^T \mathbf{\Omega} \mathbf{V} \in \mathbb{R}^{2\times 3}.
\]
Since $ \mathbf{\Omega} \in \mathrm{St}(2,3)$, one has
\begin{equation}
    \begin{aligned}
        & \displaystyle \mathbf{B}\mathbf{B}^T = (\mathbf{U}^T \mathbf{\Omega} \mathbf{V})(\mathbf{U}^T \mathbf{\Omega} \mathbf{V})^T \\
        & \displaystyle ~~~~~~~= \mathbf{U}^T \mathbf{\Omega} \mathbf{V} \mathbf{V}^T \mathbf{\Omega}^T \mathbf{U} \\
        & \displaystyle ~~~~~~~=  \mathbf{U}^T(\mathbf{\Omega}\mathbf{\Omega}^T)\mathbf{U}\\
        & \displaystyle ~~~~~~~= \mathbf{U}^T(\mathbf{I}_2)\mathbf{U}\\
        & \displaystyle ~~~~~~~= \mathbf{I}_2
    \end{aligned}
\end{equation}
i.e., $\mathbf{B} \in \mathrm{St}(2,3)$ as well.
The objective \eqref{eq:trace-max} reduces to
\[
\mathbf{Tr}(\mathbf{\Omega} \mathbf{A}^T) = \mathbf{Tr}(\mathbf{B} \mathbf{\Sigma}^T) 
= \sigma_1 B_{11} + \sigma_2 B_{22},
\]
since $\mathbf{\Sigma}^T$ has entries $(\sigma_1,\sigma_2)$ in positions $(1,1)$ and $(2,2)$ and zeros elsewhere.
The constraints $\mathbf{B}\mathbf{B}^T = \mathbf{I}_2$ imply that the two rows of $
\mathbf{B}$ are orthonormal in $\mathbb{R}^3$; in particular, $|B_{11}|\le 1$ and $|B_{22}|\le 1$ with orthogonality between rows. Since $\sigma_1\ge \sigma_2\ge 0$, the maximal value $\sigma_1+\sigma_2$ is achieved by taking
\[
\mathbf{B}^\star = \big[\mathbf{I}_2  ~~~\mathbf{0}_{2 \times 1} \big],
\]
for which $\mathbf{B}^\star(\mathbf{B}^\star)^T = \mathbf{I}_2$ and $\mathrm{tr}(\mathbf{B}^\star\mathbf{\Sigma}^T) = \sigma_1 + \sigma_2$. Thus, pulling back via $ \mathbf{\Omega} = \mathbf{U} \mathbf{B} \mathbf{V}^T$,
\begin{equation}
\label{eq:Omegastar-form}
\mathbf{\Omega}^\star =  \mathbf{U} \big[ \mathbf{I}_2  ~ \mathbf{0} \big] \mathbf{V}^T
\end{equation}
This is the unique maximizer when $\sigma_1> \sigma_2>0$ (full row rank 2). In the rank–1 or isotropic case there is a low–dimensional family of maximizers yet it will still be  captured by the polar factor detailed below.

\subsection{Equivalence to the row polar factor}
Here, we show that the formulation Eq.\eqref{eq:per_frame_rot} proposed in the main paper to obtain the closest orthonormal $2 \times 3$ $\mathbf{R}_i$ is optimal and well-grounded theoretically. Assume the row–polar decomposition of a full–row–rank say $ \mathbf{A} \in \mathbb{R}^{2 \times 3}$, then we can write 
\begin{equation}
\begin{aligned}
        & \displaystyle \mathbf{A}  = \mathbf{H}\mathbf{\Omega},  \\
        & \displaystyle\text{where}, ~\mathbf{\Omega}\in\mathrm{St}(2,3), ~\text{and}, ~~\mathbf{H} =  (\mathbf{A}\mathbf{A}^T)^{\frac{1}{2}} \in \mathbb{R}^{2 \times 2},
\end{aligned}
\end{equation}
and the Stiefel factor is given explicitly by

\begin{equation}
\label{eq:polar-factor}
\mathbf{\Omega} = \mathbf{H}^{-1} \mathbf{A} = (\mathbf{A}\mathbf{A}^T)^{-\frac{1}{2}}\mathbf{A}.
\end{equation}
To show this is indeed the optimal form let's first compute with the SVD of $\mathbf{A}\mathbf{A}^T$
\begin{equation}
    \begin{aligned}
        & \displaystyle \mathbf{A}\mathbf{A}^T = \mathbf{U} \mathbf{\Sigma} \mathbf{\Sigma}^T \mathbf{U}^T
       = \mathbf{U} \mathbf{diag}(\sigma_1^2, \sigma_2^2) \mathbf{U}^T, \\
        & \displaystyle ~~~~~\implies (\mathbf{A}\mathbf{A}^T)^{-\frac{1}{2}} = \mathbf{U} \mathbf{diag}(\sigma_1^{-1}, \sigma_2^{-1}) \mathbf{U}^T,
    \end{aligned}
\end{equation}
Now if we multiply by $\mathbf{A}$ to the above expression so that we meet the form in Eq.\eqref{eq:polar-factor}, we get
\begin{equation}
    \begin{aligned}
        & \displaystyle (\mathbf{A}\mathbf{A}^T)^{-\frac{1}{2}} \mathbf{A} = \mathbf{U} \mathbf{diag}(\sigma_1^{-1},\sigma_2^{-1}) \mathbf{U}^T (\mathbf{U} \mathbf{\Sigma} \mathbf{V}^T) \\ 
        & \displaystyle ~~~~~~~~~~~~~~~~~~~~~= \mathbf{U}\big[\mathbf{I}_2 ~~~0 \big] \mathbf{V}^T,
    \end{aligned}
\end{equation}
which matches exactly Eq.\eqref{eq:Omegastar-form}, demonstrating the proposed expression as a unique minimizer.

\formattedparagraph{Row–orthonormality.}
By construction the optimal $\mathbf{\Omega}$, symbolized as $\mathbf{\Omega}^\star$
\[
\mathbf{\Omega}^\star(\mathbf{\Omega}^\star)^T
= (\mathbf{A}\mathbf{A}^T)^{-\frac{1}{2}}\,\mathbf{A}\mathbf{A}^T(\mathbf{A}\mathbf{A}^T)^{-\frac{1}{2}}
= \mathbf{I}_2.
\]
Therefore, $\mathbf{\Omega}^\star \in \mathrm{St}(2,3)$.

\formattedparagraph{Optimality and uniqueness.}
From the trace–maximization viewpoint as defined above,
\[
\|\mathbf{A} - \mathbf{\Omega}\|_F^2 =  \|\mathbf{A}\|_F^2 + 2 - 2 \mathbf{Tr}(\mathbf{\Omega}\mathbf{A}^T),
\]
so $\mathbf{\Omega}^\star$ maximizes $ \mathbf{Tr}(\mathbf{\Omega}\mathbf{A}^T)$ and thus minimizes the cost function.
If $\mathbf{A}$ has full row rank, the solution is unique. If $\mathbf{A}$ is rank–deficient, the same formula Eq.\eqref{eq:polar-factor} holds with the inverse derived from the Moore–Penrose inverse of $(\mathbf{A} \mathbf{A}^T)^{\frac{1}{2}}$, the set of closest $\mathbf{\Omega}$ is not unique in that case yet the polar factor is one canonical choice.

\subsection{Usage in our rotation estimation}
In the rotation estimation approach, for each frame $i$ we synthesize a $2 \times 3$ block $\mathbf{\Omega}_{ik^\star}$ from the learned corrective factor associated with basis $k^\star$.
The closest valid camera block $\mathbf{R}_i$ with orthonormal rows is then obtained by the row–polar projection as :
\begin{equation}
    \begin{aligned}
        \mathbf{R}_i = \mathrm{arg} \min_{\mathbf{\Omega} \in \mathrm{St}(2,3)}\|\mathbf{\Omega}_{ik^\star} - \mathbf{\Omega} \|_F^2 = \big(\mathbf{\Omega}_{ik^\star} \mathbf{\Omega}_{ik^\star}^T\big)^{-\frac{1}{2}} \mathbf{\Omega}_{ik^\star}
    \end{aligned}
\end{equation}
which the minimizer as shown above. This projection is differentiable almost everywhere (non-smooth only when $\mathbf{\Omega}_{ik^\star}$ is rank–deficient) and extremely fast to compute, making it ideal for a \nrsfm ~rotation estimation.

\smallskip
\formattedparagraph{Geometric intuition.}
The Stiefel manifold $\mathrm{St}(2,3)$ is the set of all 2–frames in $\mathbb{R}^{3}$, where a 2-frame consists of two orthonormal row vectors. The polar decomposition for rectangular matrices finds the closest point on the Stiefel manifold to a given matrix, in the Frobenius norm. This decomposition is the rectangular analogue of the well-known polar decomposition for square matrices.

\section{Complexity Analysis}\label{sec:complexity}
The proposed \nrsfm ~solvers achieves its efficiency primarily by replacing optimization heavy procedures, i.e., semidefinite programming (SDP) for rotation estimation and alternating-direction methods (ADMM) for shape estimation, with feed-forward modules composed only of basic linear algebra primitives (matrix multiplications and SVDs). In this section, we analyze the computational and memory complexity of both stages and contrast them with the classical baselines.

\subsection{Rotation Estimation}
Let $\mathbf{W} \in \mathbb{R}^{2F \times P}$ be the measurement matrix with $F$ denoting the number of frames. The popular SDP-based Euclidean correction involves solving, for each basis triplet $k$, a SDP of the form
\[
\min_{{\mathrm{\Psi}}_k\succeq0}\ \mathrm{trace}({\mathrm{\Psi}}_k)~~
\textrm{s.t.} ~\mathbf{A}\mathrm{vec}(\mathrm{\Psi}_k)=0; \mathrm{rank}(\mathrm{\Psi}_k)\le3,
\]
where $A$ encodes the per-frame orthonormality constraints and $\mathrm{\Psi}_k\in\mathbb{R}^{3K\times3K}$. Even though the SDP is ``small and fixed'' in size ($(2K^2{-}K)$ constraints---refer \cite{dai2014simple}), each solve involves high time and memory complexity because originally proposed SDP solvers rely on interior-point methods whose cost scales cubically with the matrix dimension of $\mathrm{\Psi}_k$ and quadratically with the number of constraints. An interior-point SDP solver involves Cholesky or eigen-decomposition of matrices of size $3K$, and each iteration requires $O((3K)^6)$ time and $O((3K)^4)$ memory. Although generally $K \in [3, 13]$ is small, this fixed overhead becomes dominant as $F$ or $P$ grow, since the SDP must be solved once per sequence or per re-estimation. Furthermore, SDP solvers are CPU-bound and difficult to parallelize.

\smallskip
\formattedparagraph{Proposed Euclidean Correction.}
Our approach to Euclidean correction overcomes the SDP difficulties with a learnable corrective factor representation. Each basis triplet is parameterized as $\mathbf{Y}_k\in\mathbb{R}^{3K\times3}$ (\texttt{nn.Parameter} in PyTorch), and the associated Gram matrix $\mathrm{\Psi}_k = \mathbf{Y}_k \mathbf{Y}_k^T$ automatically satisfies the positive semidefinite (PSD) and at most rank-3 constraints by construction. For each $k$, we compute
$\mathrm{\hat{\Pi}} \mathbf{Y}_k \in \mathbb{R}^{2F \times 3}$, split this matrix into $F$ double rows
$\{\mathbf{\Omega}_{ik} = [\hat{\pi}_{2i-1} \mathbf{Y}_k;\,\hat{\pi}_{2i} \mathbf{Y}_k]\}_{i=1}^F$, and evaluate the orthogonality residuals
\[
\varepsilon_{ik}^{(1)} = (\|\mathbf{g}_{1,ik}\|_2^2-\|\mathbf{g}_{2,ik}\|_2^2)^2;~~
\varepsilon_{ik}^{(2)} = (\mathbf{g}_{1,ik} \cdot \mathbf{g}_{2,ik})^2,
\]
which constitute the orthogonality loss 
\begin{equation}
\mathcal{L}_{\text{orth}}=\frac{1}{FK}\sum_{i,k}(\varepsilon_{ik}^{(1)}+\varepsilon_{ik}^{(2)}).
\end{equation}
Backpropagation through $\mathrm{\hat{\Pi}} \mathbf{Y}_k$ updates $\mathbf{Y}_k$ via simple general matrix–matrix multiplication (GEMM).

\smallskip
\formattedparagraph{Per-iteration complexity.}
The dominant operation is the matrix multiplication $\mathrm{\hat{\Pi}} \mathbf{Y}_k$ for each $k$, whose cost is
\[
O\big((2F)(3K)\cdot3\big)=O(18FK)\ \text{ flops}.
\]
Since there are $K$ such bases, the overall cost per forward or backward pass is $ O(FK^2)\ \text{time and} ~O(FK+K^2) ~\text{memory}$. This cost is quadratic in $K$ with a very small constant factor (tens of floating-point operations per entry). The per-frame row-polar projection $\mathbf{R}_i=(\mathbf{\Omega}_{ik} \mathbf{\Omega}_{ik}^T)^{-1/2}\mathbf{\Omega}_{ik}$ is $O(F)$ and negligible. All operations are differentiable and parallelizable over $k$ and $i$.

\smallskip
\formattedparagraph{Comparison to SDP Modeling.}
Interior-point SDP solvers can scale poorly with the matrix dimension. In practice, even for small $K$, their fixed overhead and CPU-bound implementation dominate the rotation stage. While the classical SDP has superlinear dependence\footnote{A superlinear function grows more rapidly than a linear function for large values of its input.} on $K$ and is bound by dense CPU linear algebra, our rotation solver is dominated by a few GEMMs of size $(2F\times3K) \times (3K\times3)$, which execute efficiently on modern accelerators. Theoretical runtime improves from $O(K^6)$ per sequence to $O(FK^2)$ and the memory footprint from $O(K^4)$ to $O(FK)$. Empirically, on standard sequences ($F \approx 300,K \le 13$) the rotation recovered using our solver is order of magnitude faster while yielding equivalent orthonormality accuracy.

\subsection{Shape Estimation}\label{sec:complexity-shape}

\paragraph{Classical ADMM-based solvers.}
Given $ \mathbf{R}$, the shape $ \mathbf{S}$ is recovered by minimizing
\begin{equation}
\label{eq:shape-obj-complexity}
\min_\mathbf{S} \ 0.5 \|\mathbf{M} \odot(\mathbf{W} - \mathbf{R} \mathbf{S})\|_F^2 + \|\mathbf{S}^\sharp\|_{*,w}.
\end{equation}
To solve the above optimization via ADMM requires introduction of an auxiliary variable $\mathbf{U} = \mathbf{S}^\sharp$ and a dual variable $\mathbf{L}$ yielding the following iterations:
\begin{equation}
    \begin{aligned}
        & \displaystyle \mathbf{S}^{k+1} := \arg\min_S \|\mathbf{M} \odot(\mathbf{W} - \mathbf{R} \mathbf{S})\|_F^2 ~~+ \\ 
        & ~~~~~~~~~~~~~~\displaystyle \tfrac{\rho}{2}\|g(\mathbf{S})-\mathbf{U}^k + \mathbf{L}^k/\rho\|_F^2
    \end{aligned}
\end{equation}
\begin{equation}
    \begin{aligned}
        \mathbf{U}^{k+1} &:= \arg\min_\mathbf{U} \| \mathbf{U} \|_{*,w} + \tfrac{\rho}{2}\| \mathbf{U} - g(\mathbf{S}^{k+1}) - \mathbf{L}^k/\rho\|_F^2
    \end{aligned}
\end{equation}
\begin{equation}
    \begin{aligned}
       \mathbf{L}^{k+1} &:= \mathbf{L}^k + \rho(g(\mathbf{S}^{k+1}) - \mathbf{U}^{k+1}).
    \end{aligned}
\end{equation}
The $\mathbf{S}$ update rrequires solving a large linear system $(\mathbf{R}^T \mathbf{R} + \rho \mathbf{I}) \mathbf{S} = \mathbf{R}^T \mathbf{W} + \rho(g^{-1}(\mathbf{U})-\mathbf{Y}/\rho)$, which for a dense $\mathbf{R}$ costs $O((3F)^3P)$ or, at best, $O(FP)$ if block-diagonal structure is exploited. The $\mathbf{U}$ update involves a full SVD of an $F \times 3P$ matrix $O(F(3P)^2)$, and the dual update is linear.
With $N_{\text{ADMM}}$ iterations, total runtime scales as
\[
O\Big(N_{\text{ADMM}}\big(F(3P)^2+(3F)^3P\big)\Big),
\]
and memory $O(FP)$ for the primal variables plus $O(FP)$ for the duals.

\smallskip
\formattedparagraph{Proposed solver for shape estimation.}
Our unrolled layer eliminates the dual variable and linear system.
Each layer $k=0, \dots, N_k-1$ performs: \textbf{\textit{(i)}} gradient of the reprojection term: $\mathbf{G} = \mathbf{R}^T (\mathbf{M} \odot(\mathbf{R}\mathbf{S}-\mathbf{W}))$, cost $O(FP)$, \textbf{\textit{(ii)}} reshuffled shape domain update: $\mathbf{Z}^\sharp = g(\mathbf{S})-\alpha g(\mathbf{G})$, \textbf{\textit{(iii)}} WSVT: $\mathbf{X}^\sharp = \mathcal{S}_{\tau_k,w}(\mathbf{Z}^\sharp)$, cost $O(\min(F,3P)^2 \max(F,3P))$, and \textbf{\textit{(iv)}} reshape back: $ \mathbf{S} \leftarrow g^{-1}(\mathbf{X}^\sharp)$.

For a general sparse \nrsfm ~case $F > 3P$, i.e., many frames and moderate points, and so SVD dominates with cost $O(FP^2)$ per layer. Thus per-layer complexity is ${O(FP^2)\ \text{time and}\ O(FP)\ \text{memory}.}$ In dense settings where $3P \gg F$, the cost is $O(F^2P)$. Thus, the total shape-stage cost is
\(O(N_kFP^2)\) when \(F>3P\), and
\(O(N_kF^2P)\) when \(3P>F\). FISTA acceleration and continuation scheduling add only $O(FP)$ bookkeeping.

All in all, if we compare our approach to ADMM based optimization then (a) ADMM's linear solves, i.e., $(\mathbf{R}^T \mathbf{R} + \rho \mathbf{I})^{-1}$ inversion is the major bottleneck, both computationally and numerically, is removed entirely, (b) Dual variables are absent in our formulation, significantly reducing the memory footprint, (c)  our deterministic solver typically converges in $N_k\approx 40$-$60$ layers, whereas ADMM requires hundreds of iterations with penalty tuning, and (d)  each layer uses only GEMM and SVD, which are the most optimized primitives in BLAS and GPU libraries.

Asymptotically, both ADMM and our solver require SVDs of the same shape matrix, yet ADMM incurs additional $O((3F)^3P)$ linear solves per iteration, making it $1$-$2$ orders of magnitude slower. Empirically, our unrolled solver achieves identical objective values with less runtime and lower memory overhead, specifically for dense datasets.

\begin{table*}
\centering
\resizebox{1.0\linewidth}{!}{
\begin{tabular}{lcc}
\toprule
\textbf{Stage} & \textbf{Classical (SDP/ADMM)} & \textbf{Ours} \\
\midrule
Rotation estimation & $O(K^6)$ per sequence (SDP) & $O(FK^2)$ \\
Shape estimation (per iter.) & $O(F(3P)^2)+(3F)^3P$ & $O(FP^2) ~\mathrm{if} ~F > 3P, ~\mathrm{and} ~O(F^2P) ~\mathrm{if} ~3P > F $ \\
Linear solves / duals & required & none \\
Hardware acceleration & NA (CPU-bound) & excellent \\
Scales with frames $F$ & superlinear & Regime-dependent: linear in $F$, if $F > 3P$ else quadratic   \\
Scales with points $P$ & quadratic & Regime-dependent: linear in $P$, if $3P > F$ else quadratic \\
Scales with bases $K$ & superlinear ($K^6$) & quadratic ($K^2$) \\
\bottomrule
\end{tabular}
}
\caption{Asymptotic complexity comparison between SOTA SDP and ADMM-based \nrsfm ~and the proposed approach.}\label{tab:complexity-summary}
\end{table*}

\subsection{Discussion} \label{sec:complexity-summary}
For typical benchmarks ($F \approx 300$, $P \approx 30-100$, $K \le 13$), the proposed approach runs in a few seconds on NVIDIA A6000 GPU or Apple M3 processor, compared to several minutes for SDP/ADMM pipelines on CPU/GPU's. Our introduced improvement transforms \nrsfm ~matrix factorization approach from a slow, research-oriented method into a practical and scalable tool for non-rigid deformable object 3D reconstruction so that it can further be used in downstream computer vision tasks which at present is mostly machine-learning driven. Table~\ref{tab:complexity-summary} summarizes the asymptotic complexity comparison of both SDP/ADMM based SOTA methods with our proposed \nrsfm ~approach.

\section{Lipschitz properties: Re-projection term}\label{sec:lipschitz}
In our formulation, the overall objective is composed of $f(\mathbf{S})$ and $g(\mathbf{S})$, where  
\[
f(\mathbf{S}) =  0.5 * \| \mathbf{M} \odot ( \mathbf{R}\mathbf{S} - \mathbf{W}) \big\|_F^2, ~~\textrm{with} ~\mathbf{S} \in \mathbb{R}^{3F\times P},
\]
It enforces re-projection consistency between the estimated 3D structure $\mathbf{S}$ and the observed 2D points $\mathbf{W}$. In our formulation, $g(\mathbf{S})$ is the regularization, which is the non-smooth part that promotes the low-rank structure across frames. Now to apply proximal gradient method, we need to take gradient step on $f()$ and the proximal step on $g()$. For that gradient step to be valid and stable, the gradient of $f()$ must be Lipschitz continuous for some constant $L > 0$ (Lipschitz constant). This condition tells us that the gradient of $f$ does not change too abruptly and geometrically the curvature of $f$ is bounded.

Next, we will show step by step that the smooth part of the our \nrsfm ~objective is \textit{\textbf{(i)}} continuously differentiable with an explicit gradient, and \textit{\textbf{(ii)}} has a globally Lipschitz-continuous gradient, i.e., $f$ is $L$--smooth with
\[
L \le \|\mathbf{R}\|_2^2,
\]
where, $\| \cdot\|_2$ denotes the spectral norm and under orthography $\|\mathbf{R}\|_2 \le 1$.

\smallskip
\subsection{Preliminaries and notations}

\noindent
\textbf{Hilbert space}: Assume V, a vector space that is metrically \textit{complete} with respect to the metric induced by its own norm. If V's norm induces a complete metric space and is itself induced by an inner product on V, we say V is a Hilbert space \cite{adler2021hilbert, young1988introduction}. Here, the notion of Hilbert space  $\mathbb{R}^{m \times n}$ is used with the Frobenius inner product $\langle \textbf{X}, \textbf{Y} \rangle = \mathbf{Tr}(\mathbf{X}^T \mathbf{Y})$ and the induced Frobenius norm $\| \mathbf{X} \|_F = \sqrt{\langle \mathbf{X}, \mathbf{X}\rangle}$.

As defined in the main paper, let $ \mathbf{R} \in \mathbb{R}^{2F \times 3F}$ be the block-diagonal rotation matrix, $\mathbf{M} \in \{0, 1\}^{2F \times P}$ a fixed binary visibility mask and $\mathbf{W} \in \mathbb{R}^{2F \times P}$ the measurement matrix. Let's define the linear operators
\[
\mathcal{H}_\mathbf{M}(\mathbf{Z}) \triangleq \mathbf{M} \odot \mathbf{Z} ~~~~\text{(Hadamard masking)}, ~~\mathcal{R}(\mathbf{S}) \triangleq \mathbf{R} \mathbf{S}.
\]
We will use of the following facts: \textit{\textbf{(i)}} $\mathcal{H}_\textbf{M}$ is an orthogonal projection: $\mathcal{H}_\textbf{M} = \mathcal{H}_\textbf{M}^\ast = \mathcal{H}_\textbf{M}^2$ and
$\|\mathcal{H}_\mathbf{M}(\mathbf{Z})\|_F \le \| \mathbf{Z} \|_F$ (i.e., $\|\mathcal{H}_M\|_{op}=1$).
\textit{\textbf{(ii)}} $\mathcal{R}$ is bounded with $\|\mathcal{R}\|_{op}=\| \mathbf{R} \|_2$, and $\mathcal{R}^\ast(\mathbf{T}) = \mathbf{R}^T \mathbf{T}$. For convenience, lets define the composite linear operator
\[
\mathcal{A}: \mathbb{R}^{3F \times P} \to \mathbb{R}^{2F \times P}, ~~~~
\mathcal{A}(\mathbf{S}) = \mathcal{H}_\mathbf{M}(\mathcal{R}(\mathbf{S})) = \mathbf{M} \odot( \mathbf{R}\mathbf{S}),
\]
and its adjoint $\mathcal{A}^\ast(\mathbf{T}) = \mathcal{R}^\ast(\mathcal{H}_\mathbf{M}(\mathbf{T})) = \mathbf{R}^T(\mathbf{M} \odot \mathbf{T})$. Using \textit{\textbf{(i)}} and \textit{\textbf{(ii)}},

\begin{equation}\label{eq:A-norm}
\|\mathcal{A}\|_{op} \le \|\mathcal{H}_\mathbf{M}\|_{op} \|\mathcal{R}\|_{op} = \|\mathbf{R}\|_2.
\end{equation}

\subsection{Differentiability and the gradient of $f$}
We rewrite $f$ in operator form:
\[
f(\mathbf{S}) = 0.5*\|\mathcal{A}(\mathbf{S}) - \mathbf{B} \|_F^2, ~~~~ \mathbf{B} \triangleq \mathcal{H}_\textbf{M}(\mathbf{W}) = \mathbf{M} \odot \mathbf{W}.
\]
Using standard rules for least squares in Hilbert spaces\footnote{Linear least squares in a Hilbert space extends the concept of minimizing the squared error to infinite-dimensional vector spaces, where the solution is found by finding a vector that is orthogonal to the error space.},
\[
\nabla f(\mathbf{S}) = \mathcal{A}^\ast\big(\mathcal{A}(\mathbf{S})- \mathbf{B} \big) = \mathbf{R}^T \Big(\mathbf{M} \odot \big( \mathbf{R} \mathbf{S} - \mathbf{W} \big)\Big).
\]
Thus $f$ is (globally) $\mathcal{C}^1$ with affine gradient $\nabla f(\mathbf{S}) = \mathcal{A}^\ast \mathcal{A} (\mathbf{S}) - \mathcal{A}^\ast \mathbf{B}$.

\subsection{Lipschitz continuity of the gradient ($L$--smoothness)}
We say $f$ satisfies the Lipschitz condition \cite{lipschitz1877lehrbuch} on a set $D$ if there is a constant $L \geq 0$
such that \[ |f(t, u) - f(t, v)| \leq L |u-v|  ~~\forall (t, u), (t, v) \in D.  \] We now show that $\nabla f$ is globally Lipschitz and compute an admissible constant.
For any $\mathbf{S}_1, \mathbf{S}_2\in\mathbb{R}^{3F \times P}$,
\begin{align*}
\nabla f(\mathbf{S}_1) - \nabla f(\mathbf{S}_2) & = \mathcal{A}^\ast\big(\mathcal{A}(\mathbf{S}_1)- \mathbf{B} \big) - \mathcal{A}^\ast\big(\mathcal{A}(\mathbf{S}_2)- \mathbf{B} \big)\\
& = \mathcal{A}^\ast \mathcal{A}\,(\mathbf{S}_1 - \mathbf{S}_2).
\end{align*}
Taking norms and using sub-multiplicativity of the operator norm,
\begin{equation}\label{eq:smooth-bound}
\begin{aligned}
    & \displaystyle \|\nabla f(\mathbf{S}_1) - \nabla f(\mathbf{S}_2)\|_F \le \|\mathcal{A}^\ast\|_{op} \|\mathcal{A}\|_{op} \|\mathbf{S}_1 - \mathbf{S}_2\|_F \\  & \displaystyle  ~~~~~~~~~~~~~~~~~~~~~~~~~~~~~~~~~~~~~=\|\mathcal{A}\|_{op}^2\,\|\mathbf{S}_1 - \mathbf{S}_2\|_F,
\end{aligned}
\end{equation}
since $\|\mathcal{A}^\ast\|_{op}=\|\mathcal{A}\|_{op}$ in Hilbert spaces.
Combining \eqref{eq:smooth-bound} with \eqref{eq:A-norm}, we obtain the global Lipschitz bound
\[
\|\nabla f(\mathbf{S}_1)-\nabla f(\mathbf{S}_2)\|_F\ \le\ \| \mathbf{R} \|_2^2\,\|\mathbf{S}_1-\mathbf{S}_2\|_F, ~\forall ~\mathbf{S}_1, \mathbf{S}_2,
\]
i.e., $f$ is $L$--smooth with any $L \ge \|\mathbf{R} \|_2^2$.
Under orthographic projection, $R$ has orthonormal rows blockwise and $\|\mathbf{R}\|_2 \le 1$, hence one may take $L = 1$. The above single inequality is the Lipschitz-gradient proof.

\smallskip
\formattedparagraph{Alternative proof (direct).}
Without operator notation, let $\Delta \mathbf{S} = \mathbf{S}_1 - \mathbf{S}_2$. Then
\[
\nabla f(\mathbf{S}_1)-\nabla f(\mathbf{S}_2) = \mathbf{R}^T \big(\mathbf{M} \odot( \mathbf{R} \Delta \mathbf{S})\big).
\]
Using the contractivity $\| \mathbf{M} \odot \mathbf{X} \|_F\le\| \mathbf{X} \|_F$ for Hadamard masking and sub-multiplicativity,
\begin{equation}
    \begin{aligned}
        & \displaystyle \|\nabla f(\mathbf{S}_1)-\nabla f(\mathbf{S}_2)\|_F \le \| \mathbf{R}^T \|_2 \|\mathbf{M} \odot( \mathbf{R} \Delta \mathbf{S})\|_F \\ 
        & \displaystyle  ~~~~~~~~~~~~~~~~~~~~~~~~~~~~~~~~~~~~~\le \|\mathbf{R}\|_2 \|\mathbf{R} \Delta \mathbf{S}\|_F \le \|\mathbf{R}\|_2^2 \|\Delta \mathbf{S}\|_F,
    \end{aligned}
\end{equation}
which is the same bound.

\subsection{On Lipschitz continuity of $f$ itself}
A globally Lipschitz function $h$ satisfies $|h(\mathbf{X})-h(\mathbf{Y})|\le K \| \mathbf{X}-\mathbf{Y} \|$ for some $K$ independent of $\mathbf{X}, \mathbf{Y}$. Quadratic functions like $f(\mathbf{S})=\frac{1}{2}\|\mathcal{A}(\mathbf{S})- \mathbf{B}\|_F^2$ are in general not globally Lipschitz on an unbounded domain as they grow quadratically, but they are Lipschitz on any bounded set. For our case $\mathbf{S}_1, \mathbf{S}_2$ can safely be assumed in a finite ball $\{\mathbf{S}: \|\mathbf{S}\|_F \le r ~\textrm{(radius)} \}$, the mean value theorem in Hilbert spaces gives

\begin{equation}
    \begin{aligned}
        & \displaystyle |f(\mathbf{S}_1)-f(\mathbf{S}_2)| \\
        & \displaystyle \le \sup_{t\in[0,1]} \|\nabla f(\mathbf{S}_2 + t(\mathbf{S}_1 - \mathbf{S}_2))\|_F\ \|\mathbf{S}_1-\mathbf{S}_2\|_F \\
        & \displaystyle \le \Big(\|\mathcal{A}\|_{op}^2 r + \|\mathcal{A}\|_{op}\,\|\mathbf{B}\|_F\Big) \|\mathbf{S}_1- \mathbf{S}_2\|_F
    \end{aligned}
\end{equation}
so $f$ is Lipschitz on bounded sets with a constant that depends linearly on the radius $r$.
For algorithmic purposes (step-size selection, convergence of proximal-gradient), the relevant property is the Lipschitz continuity of the gradient established above.

\subsection{Consequences (step sizes and descent)}
Since $f$ has an $L$--Lipschitz gradient with $L\le \|\mathbf{R}\|_2^2$, any fixed step $\alpha\in(0,1/L]$ guarantees that the standard descent lemma holds:
\[
f(\mathbf{S} + \Delta)\ \le\ f(\mathbf{S})  +  \langle \nabla f(\mathbf{S}),\Delta\rangle\ + \tfrac{L}{2} \|\Delta\|_F^2, ~~\forall ~\mathbf{S},\Delta.
\]
This is the condition needed to ensure monotonic decrease in the proximal-gradient (and FISTA-with-restart \cite{xiang2021fista}) iterations used in our sharp-domain spectral–proximal solver. In particular, under orthography one may safely choose $\alpha\le 1$.

\smallskip
\formattedparagraph{\textit{Q. Why use the notion of Hilbert space to establish the last proof (Sec.\ref{sec:lipschitz})?}}
Such an abstract functional analysis to prove this is optional, but treating matrices as elements of a Hilbert space with the Frobenius inner product makes the proof short, clean, and modular. Furthermore, in a Hilbert space every bounded linear operator $\mathcal{A}$ has a unique adjoint $\mathcal{A}^\ast$ defined by 
\[
\langle \mathcal{A}(\mathbf{X}), \mathbf{Y} \rangle = \langle \mathbf{X}, \mathcal{A}^\ast(\mathbf{Y}) \rangle ~\forall ~\mathbf{X}, \mathbf{Y} ~\textrm{and}, \|\mathcal{A}\|_{op} = \|\mathcal{A}^\ast \|_{op}.
\]
This lets us write gradients and Lipschitz bounds using operator norms instead of componentwise calculations. Other benefits it brings which is specific to the form proposed in the \nrsfm ~formulation are: (a) Clean expression for the gradient, i.e., $\nabla f(\mathbf{S}) = \mathcal{A}^\ast\big(\mathcal{A}(\mathbf{S})- \mathbf{B} \big)$ follows from the Reisz representation in a Hilbert space \cite{adler2021hilbert}, (b) Immediate Lipschitz bound, i.e., $\|\nabla f(\mathbf{S}_1) - \nabla f(\mathbf{S}_2)\|_F = \|\mathcal{A}\|_{op}^2\,\|\mathbf{S}_1 - \mathbf{S}_2\|_F$. The only thing to upper bound is $\|\mathcal{A}\|$, which we establish by $\|\mathbf{R}\|_2$ due to $\|\mathcal{H}_\mathbf{M}\| \leq 1$, and lastly (c) if we change the mask $\mathbf{M}$ or the camera model $\mathbf{R}$ the same expression could work, we just need to update the $\|\mathcal{A}\|$ norm.

Long story short, the Hilbert-space view lets us treat \nrsfm ~re-projection map as a single bounded linear operator. After that all the usual least-squares calculus and norm inequalities apply verbatim.

\section{Additional Discussion and Results}
\noindent
\revisedone{
\textbf{1. Sequence-wise accuracy tradeoffs.} We use a sequential rotation-then-shape design to preserve the main goal of the solver. A coordinate-descent refinement that alternates between camera and shape updates may further improve accuracy on some sequences, especially when the initial camera estimate is imperfect, but it would reintroduce an outer iterative loop and increase runtime. We therefore treat such refinement as an optional accuracy-oriented extension rather than part of the fast no-cone, no-dual solver.
}

\smallskip
\noindent
\revisedone{
\textbf{2. Accuracy limit: coordinate descent.} The sequence-wise behavior of our apporach is consistent with the priors used by competing methods. Our solver performs strongly when the factor-side rotation estimate is stable and the low-rank shape-space prior is sufficient, e.g., Walking and Stretch. Larger gaps appear on sequences such as Drink, Dance, and Shark, where methods such as TSM and OPM benefit from additional temporal, Procrustean, rotation-averaging, or deformation priors. Our method deliberately avoids these priors to keep the solver fast and broadly applicable, which explains the observed accuracy-runtime tradeoff.
}

\smallskip
\noindent
\revisedone{
\formattedparagraph{3. Stage-wise runtime accounting.}
To clarify the runtime reported in the main paper, Table~\ref{tab:stage_runtime} separates the cost of the two solver stages on the CMU MoCap sequences and dense face benchmarks. The rotation time corresponds to optimizing the factor-side Euclidean correction and applying the row-polar projection to recover the block-orthographic camera matrix. The shape time corresponds to the fixed-depth no-dual spectral-proximal solver after the rotations are fixed. The total runtime is the sum of these two stages. In particular, longer, denser, and challenging sequences such as Drink, Dance, Shark, and Dense Face Sequence incur higher cost, but the total runtime remains within a practical range.
}

\smallskip
\noindent
\revisedone{
\formattedparagraph{4. Sensitivity to the shape-space rank \(K\).}
Table~\ref{tab:k_sensitivity} shows the result on the standard CMU MoCap sequences with different $K$ values. We vary \(K\in\{10,11,12,13\}\). \textit{The results show that the proposed solver is not fragile to a narrowly tuned rank value}. The overall trend remains stable and does not show a monotonic advantage from simply increasing model capacity. In several cases, nearby \(K\) values slightly improve the reported error.
}

\begin{table}[t]
\centering
\small
\begin{tabular}{lcccc}
\toprule
Dataset & Rotation time & Shape time & Total time \\
\midrule
\multicolumn{4}{c}{\textbf{CMU MoCap Sequence}}\\
\midrule
Drink & 8.51 & 6.45 & 14.96  \\
Pickup & 7.79 & 4.88 & 12.67 \\
Yoga & 7.73 & 4.19 & 11.92 \\
Stretch & 7.83 & 4.21 & 12.04 \\
Dance & 8.57 & 6.39 & 14.96 \\
Face & 8.42 & 4.66 & 13.08 \\
Walking & 8.52 & 4.84 & 13.36 \\
Shark & 8.27 & 9.51 & 17.78 \\
\midrule
\multicolumn{4}{c}{\textbf{Dense Face Sequence}}\\
\midrule
Face Seq.1 & 1.24 & 7.05 & 8.29 \\
Face Seq.2 & 1.25 & 6.93 & 8.18 \\
Face Seq.3 & 1.73 & 13.28 & 15.01 \\
Face Seq.4 & 1.65 & 13.51 & 15.16  \\
\bottomrule
\end{tabular}
\caption{\revisedone{Stage-wise runtime of our solver on sparse MoCap and dense face benchmarks. Rotation time includes factor-side Euclidean correction and row-polar projection. Shape time includes the fixed-depth no-dual spectral-proximal solver after rotations are fixed. Total time is the combined solver runtime. All times are reported in seconds.}}
\label{tab:stage_runtime}
\end{table}

\begin{table}[h]
\centering
\small
\setlength{\tabcolsep}{6pt}
\renewcommand{\arraystretch}{1.08}
\begin{tabular}{lcccc}
\toprule
\text{Sequence} & $K=10$ & $K=11$ & $K=12$ & $K=13$ \\
\midrule
Drink   & 0.0161 & \underline{0.0166} & 0.0164 & \textbf{0.0153} \\
Pickup  & 0.0272 & \textbf{0.0239} & \underline{0.0267} & 0.0256 \\
Yoga    & 0.0257 & \textbf{\underline{0.0148}} & 0.0236 & 0.0207 \\
Stretch & 0.0122 & \textbf{0.0118} & 0.0119 & \underline{0.0128} \\
\bottomrule
\end{tabular}
\caption{\revisedone{Sensitivity of $e_{3d}$ to the shape-space rank $K$ on CMU MoCap sequences. The underlined entries indicate the $K$ values used in the main-paper comparison following prior shape-space baselines, while bold entries indicate the best result in the sweep. The results show that the proposed solver is stable across nearby ranks and is not dependent on aggressive per-sequence rank tuning.}
}
\label{tab:k_sensitivity}
\end{table}

\begin{table}[t]
\resizebox{1.0\linewidth}{!}{
\begin{tabular}{|c|c|c|c|}
\hline
$\eta \downarrow$ /cap $\rightarrow$ & 0.90 & 0.95 & 0.99 \\ \hline
1.02                                 &   (0.0660, 0.0211)    &  (0.0108, 0.0200)    & (0.0112, 0.0177)     \\ \hline
1.05                                 & (0.1194, 0.0271)     &  (0.0640, 0.0265)    & (0.0168, 0.0267)     \\ \hline
1.10                                 &   (0.1369, 0.0577)   &  (0.1437, 0.0592)     &  (0.1248, 0.0591)    \\ \hline
\end{tabular}
}
\caption{\((e_{3d}^{\mathrm{Drink}},e_{3d}^{\mathrm{Pickup}})\) with $N_\textrm{pres} = 0$}\label{tab:sens_npres0}
\end{table}

\begin{table}[t]
\resizebox{1.0\linewidth}{!}{
\begin{tabular}{|c|c|c|c|}
\hline
$\eta \downarrow$ /cap $\rightarrow$ & 0.90 & 0.95 & 0.99 \\ \hline
1.02                                 & (0.0510, 0.0233)     &   (\textbf{0.0108}, 0.0204)   &   (0.0111, \textbf{0.0176})   \\ \hline
1.05                                 &  (0.1219, 0.0280)    &   (0.0618, 0.0264)   &   (\textcolor{blue}{0.0166}, \textcolor{blue}{0.0267})  \\ \hline
1.10                                 &  (0.1360, 0.0698)    &   (0.1411, 0.0621)    & (0.1234, 0.0590)     \\ \hline
\end{tabular}
}
\caption{\((e_{3d}^{\mathrm{Drink}},e_{3d}^{\mathrm{Pickup}})\) with $N_\textrm{pres} = 1$}\label{tab:sens_npres1}
\end{table}

\begin{table}[t]
\resizebox{1.0\linewidth}{!}{
\begin{tabular}{|c|c|c|c|}
\hline
$\eta \downarrow$ /cap $\rightarrow$ & 0.90 & 0.95 & 0.99 \\ \hline
1.02                                 & (0.0976, 0.0238)     & (0.0108, 0.0224)     &   (0.0113, 0.0186)   \\ \hline
1.05                                 &  (0.1496, 0.0261)    &  (0.1205, 0.0274)    &  (0.0249, 0.0268)    \\ \hline
1.10                                 &  (0.1690, 0.0880)    &  (0.1658, 0.0820)     &  (0.1570, 0.0826)    \\ \hline
\end{tabular}
}
\caption{\((e_{3d}^{\mathrm{Drink}},e_{3d}^{\mathrm{Pickup}})\)  with $N_\textrm{pres} = 2$}\label{tab:sens_npres2}
\end{table}

\noindent
\revisedone{
\formattedparagraph{5. Sensitivity to continuation and spectral safeguards.}
Tables~\ref{tab:sens_npres0}-\ref{tab:sens_npres2} study the effect of the continuation and spectral-safeguard parameters in the unrolled shape solver. Each entry reports \((e_{3d}^{\mathrm{Drink}},e_{3d}^{\mathrm{Pickup}})\), while varying the number of preserved singular values \(N_{\mathrm{pres}}\), the continuation rate \(\eta\), and the shrinkage cap. All other parameters are kept fixed. The default setting used in the main paper is \(N_{\mathrm{pres}}=1\), \(\eta=1.05\), and \(\mathrm{cap}=0.99\).
}

\revisedone{
The results show that the solver behavior is predictable rather than arbitrary. The continuation rate \(\eta\) has the strongest effect: a large value such as \(\eta=1.10\) relaxes the shrinkage too quickly and consistently degrades reconstruction accuracy. Slower or moderate continuation, \(\eta\in\{1.02,1.05\}\), is substantially more stable. The number of preserved modes also has the expected effect, i.e., preserving no mode or only the dominant mode gives similar performance in stable regimes, while preserving two modes can weaken low-rank regularization and increase error, especially on Drink. Finally, caps in the range \(0.95\)-\(0.99\) are more reliable than \(0.90\), suggesting that overly restrictive shrinkage caps can prevent effective spectral regularization.}

\revisedone{
\textbf{Importantly, the main-paper default setting on these two sequences gives stable results.} Yet, aggressively striving for best possible $(\eta, \textrm{cap})$ may further improve results. For instance, some \(\eta=1.02\) settings yield slightly lower errors. \textit{Thus, the reported main results are not obtained by aggressive per-sequence tuning}. Instead, \textbf{the default parameters are used as a single fixed setting across experiments, providing a stable accuracy-runtime tradeoff while preserving the intended no-cone no-dual solver design.}
}

\begin{table}[t]
\centering
\small
\begin{tabular}{lccc}
\toprule
Sequence & Ours \(\mathbf{R}\) & GT \(\mathbf{R}\) & $\textrm{Gap}$  \\
\midrule
Drink   & 0.0166 & 0.0158 & 0.0008 \\
Pickup  & 0.0267 & 0.0237 & 0.0030 \\
Yoga    & 0.0148 & 0.0148 & 0.0000 \\
Stretch & 0.0127 & 0.0128 & 0.0001 \\
\bottomrule
\end{tabular}
\caption{
Pose-impact diagnostic on MoCap sequences with ground-truth rotations. We compare the 3d reconstruction results using estimated rotation and ground-truth rotation, respectively. The gap is computed as $ \textrm{Gap} = |e_{3d}(\mathrm{Ours}~\mathbf{R})-e_{3d}(\mathrm{GT}~\mathbf{R})|$.
}
\label{tab:oracle_pose}
\end{table}

\smallskip
\noindent
\revisedone{
\formattedparagraph{6. Pose-impact diagnostic using GT rotations.}
To isolate the influence of camera-pose estimation on final shape recovery, Table~\ref{tab:oracle_pose} compares the proposed shape solver using our estimated rotations against the same solver using ground-truth rotations on MoCap sequences where GT rotations are available. This diagnostic follows the same spirit as prior shape-space \nrsfm ~analyses. In particular, Kumar and Van Gool~\cite{kumar2022organic} evaluated rotation accuracy using GT rotations and observed that, under the factorization model \(\mathbf{W}=\mathbf{R}\mathbf{S}\), even ground-truth rotations do not by themselves guarantee ground-truth shape recovery. Our results are consistent with this observation. Replacing our estimated rotations with GT rotations changes \(e_{3d}\) only marginally: \(8\times10^{-4}\) on Drink, \(3\times10^{-3}\) on Pickup, \(0\) on Yoga, and \(10^{-4}\) on Stretch. Thus, the factor-side Euclidean correction provides camera estimates accurate enough for downstream shape recovery, and the remaining error is largely governed by the low-rank shape prior, deformation complexity, and the intended speed-accuracy tradeoff of the no-dual solver rather than gross pose-estimation gap.
}

\begin{figure}[h]
    \centering
    \includegraphics[width=\linewidth]{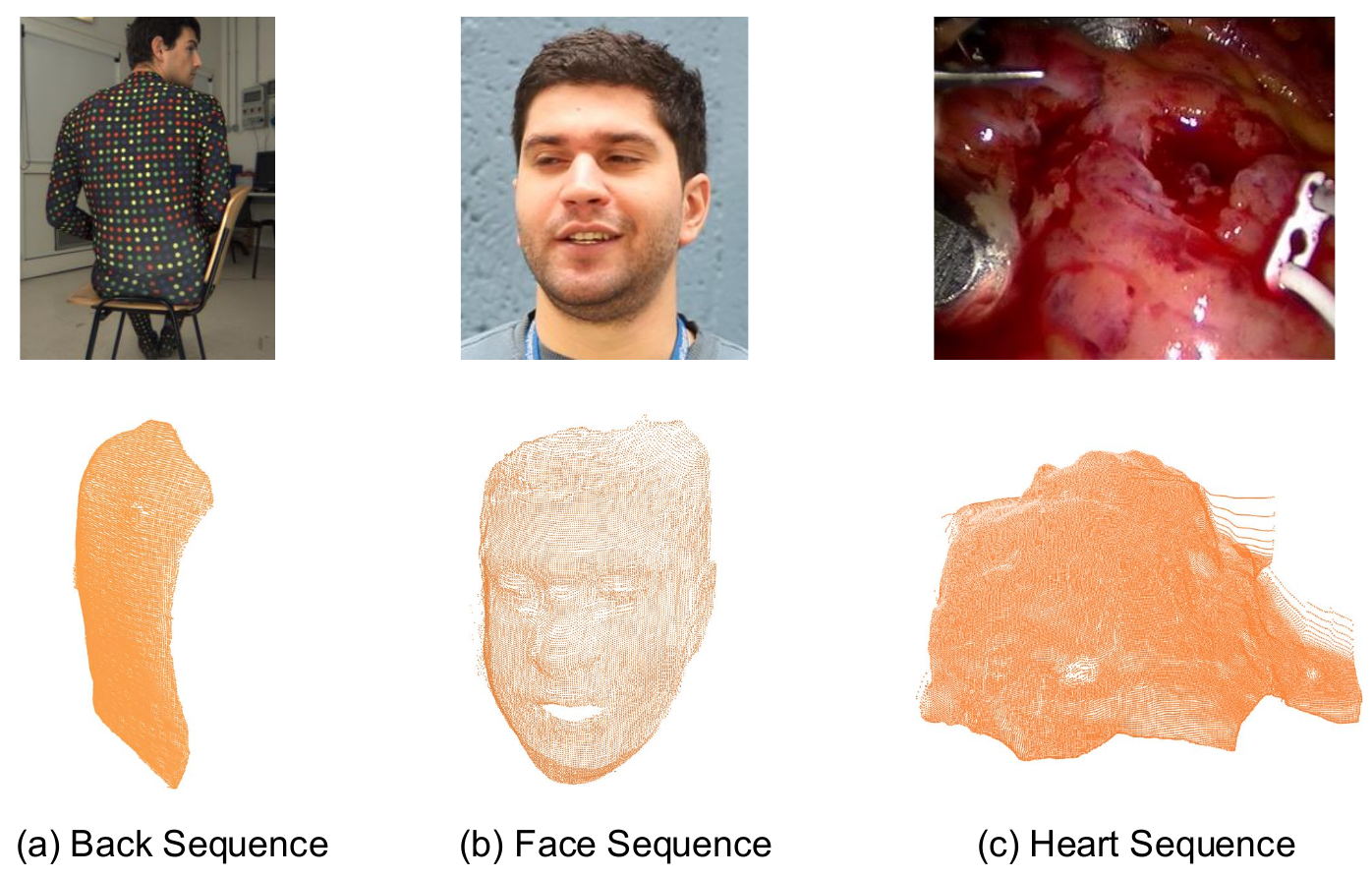}
    \caption{Qualitative results on real dense dataset sequence \cite{garg2013dense}.}
    \label{fig:real-dense-seq-results}
\end{figure}

\smallskip
\noindent
\revisedone{
\formattedparagraph{7. Qualitative evaluation on real dense sequences.}
We further evaluate our approach on the real dense \nrsfm ~dataset \cite{garg2013dense,garg2013variational}. This dataset consists of monocular videos captured in natural settings with varying illumination and large non-rigid deformation. It contains three sequences namely Face, Back, and Heart, with $120$, $150$, and $80$ frames, respectively. The dataset provides dense 2D feature tracks for all three sequences, with $28,332$ tracks for Face, $20,561$ tracks for Back, and $68,295$ tracks for Heart without GT 3D shape. Figure ~\ref{fig:real-dense-seq-results} shows using our approach, demonstrating visually coherent dense 3D structure from real monocular videos.
}

\begin{figure*}[t]
    \centering
    \includegraphics[width=1.0\linewidth]{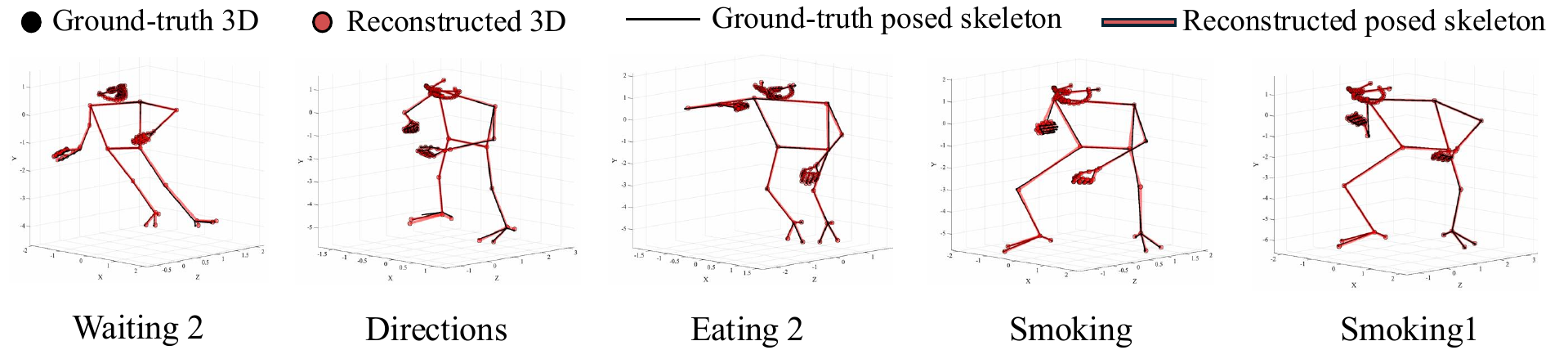}
    \caption{Qualitative results on different categories present in H3WB dataset \cite{shi2024non}.}
    \label{fig:H3WB-Visual-results}
\end{figure*}

\begin{figure*}[t]
    \centering
    \includegraphics[width=1.0\linewidth]{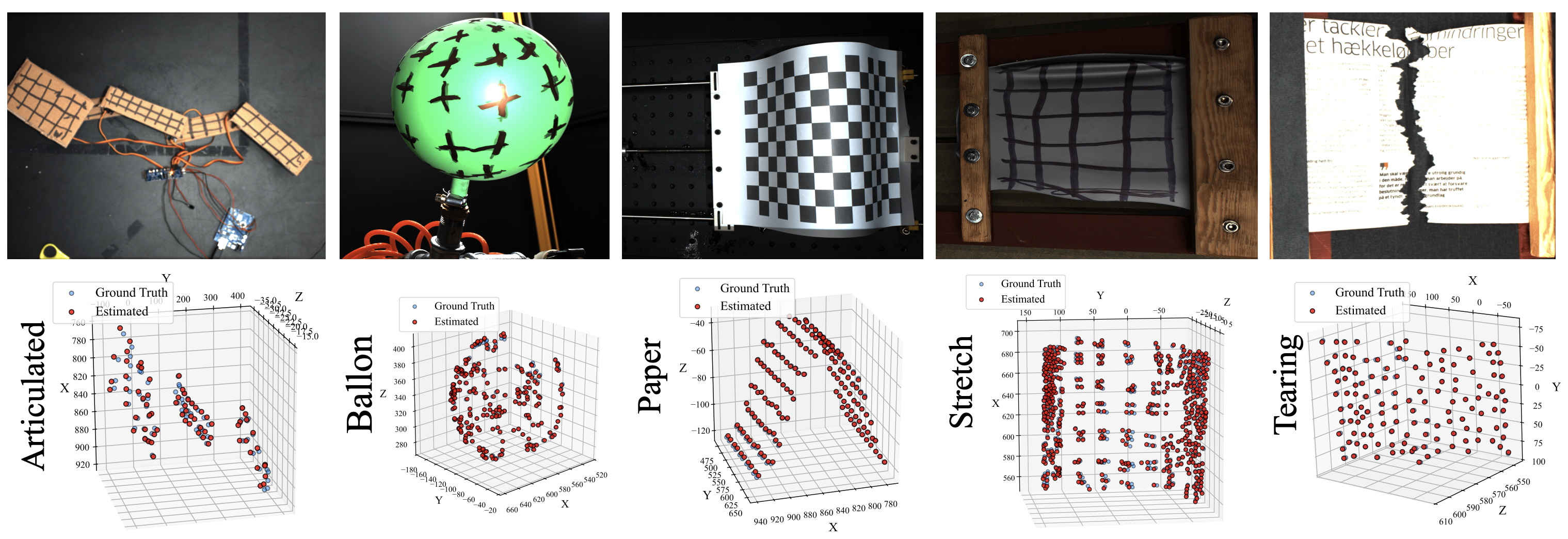}
    \caption{Qualitative results on different categories in \nrsfm ~challenge dataset \cite{jensen2018benchmark}.}
    \label{fig:Nrsfmchallenge-Visual-results}
\end{figure*}

\begin{figure*}[t]
    \centering
    \includegraphics[width=1.0\linewidth]{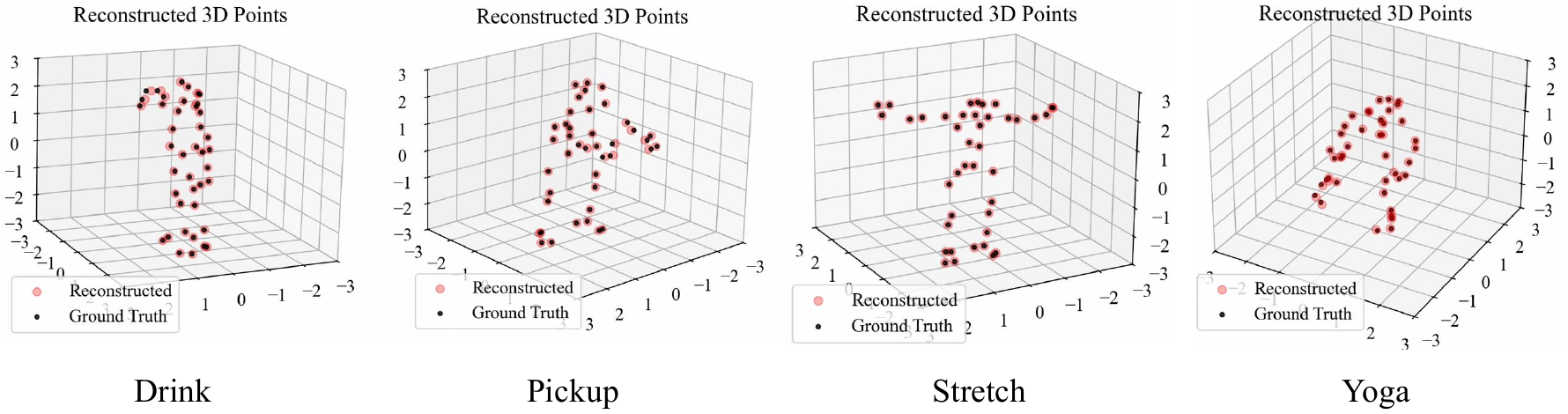}
    \caption{Qualitative results on CMU Mocap dataset \cite{akhter2009nonrigid}.}
    \label{fig:cmumocap-Visual-results}
\end{figure*}

\begin{figure*}[t]
    \centering
    \includegraphics[width=1.0\linewidth]{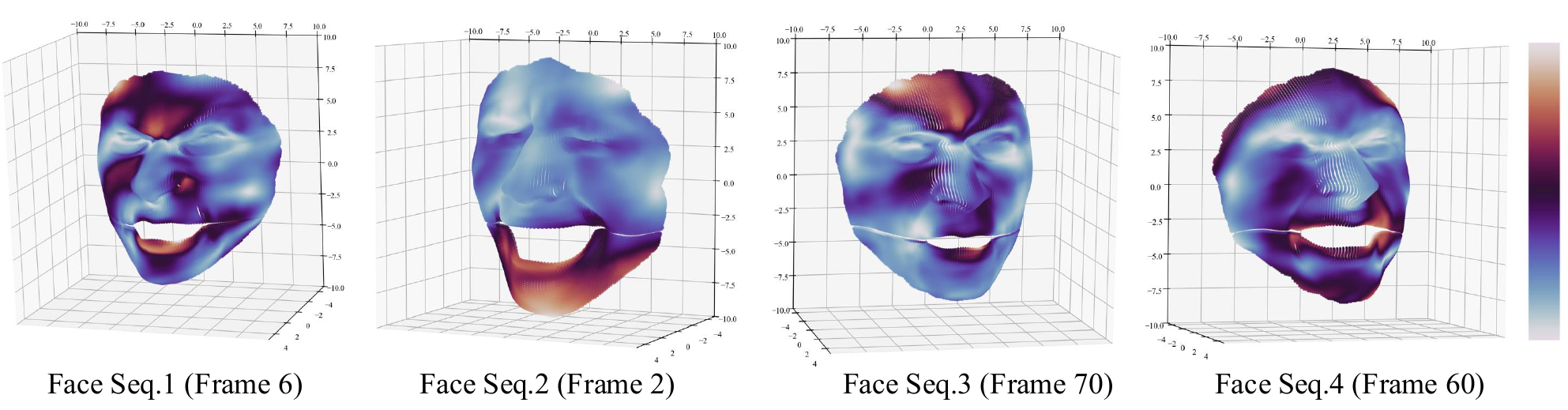}
    \caption{Qualitative results on Dense \nrsfm ~Face dataset sequence \cite{garg2013dense} showing the $1^\textrm{st}$ order deformation field.}
    \label{fig:denseface-Visual-results}
\end{figure*}

\smallskip
\noindent
\formattedparagraph{Additional Visualization.} Figure \ref{fig:H3WB-Visual-results}, Figure \ref{fig:Nrsfmchallenge-Visual-results}, Figure \ref{fig:cmumocap-Visual-results}, and Figure \ref{fig:denseface-Visual-results} provide additional qualitative results on H3WB \cite{shi2024non}, \nrsfm ~challenge benchmark \cite{jensen2018benchmark}, CMU Mocap \cite{akhter2009nonrigid}, and Dense Face Dataset \cite{garg2013dense}, respectively.

\end{document}